\documentclass[11pt]{article}
\usepackage[hyperref]{naacl2021}

\usepackage{times}
\usepackage{latexsym}
\usepackage{multirow}
\usepackage{booktabs}
\usepackage{comment}
\usepackage{graphicx}
\usepackage{adjustbox}
\usepackage{todonotes}
\usepackage{xcolor}
\usepackage{amssymb}

\usepackage{xcolor}
\usepackage{microtype}

\title{Do RNN States Encode Abstract Phonological Processes?}

\usepackage{tipa}

\newcommand{\ubc}{\textrm{\textipa{Z}}}
\newcommand{\iu}{\textrm{\textipa{Q}}}
\newcommand{\cub}{\textrm{\textipa{X}}}

\author{Miikka Silfverberg$^{\ubc}$~~~\;~~~\textbf{Francis Tyers}$^{\iu}$~~~\;~~~\textbf{Garrett Nicolai}$^{\ubc}$~~~\;~~~\textbf{Mans Hulden}$^{\cub}$ \\
$^{\ubc}$University of British Columbia~~~\;~~~$^{\iu}$Indiana University\\
$^\cub$University of Colorado\\
\texttt{first.last@ubc.ca}~~~\;~~~\texttt{ftyers@iu.edu}~~~\;~~~\texttt{first.last@colorado.edu}}
\date{}

\begin{document}
\maketitle
\begin{abstract}
Sequence-to-sequence models have delivered impressive results in word formation tasks such as morphological inflection, often learning to model subtle morphophonological details with limited training data. Despite the performance, the opacity of neural models makes it difficult to determine whether complex generalizations are learned, or whether a kind of separate rote memorization of each morphophonological process takes place. To investigate whether complex alternations are simply memorized or whether there is some level of generalization across related sound changes in a sequence-to-sequence model, we perform several experiments on Finnish consonant gradation---a complex set of sound changes triggered in some words by certain suffixes. We find that our models often---though not always---encode 17 different consonant gradation processes in a handful of dimensions in the RNN.  We also show that by scaling the activations in these dimensions we can control whether consonant gradation occurs and the direction of the gradation. %In some trained models, a single dimension entirely controls consonant gradation behavior.

\end{abstract}

\section{Introduction}

Recent work on computational morphology demonstrates that neural networks can very effectively learn to inflect words, given adequate amounts of training data \cite{cotterell2016sigmorphon,cotterell2017conll}. However, in computational morphology and in NLP at large, the interpretability of neural models remains a serious concern \cite{doshi2017towards}---it is  unclear how networks trained to inflect words actually accomplish their task. It is also unclear to which extent networks are able to learn linguistic generalizations from their input data instead of simply memorizing training examples and exhibiting a kind of nearest-neighbor behavior.

In this paper, we shed light on what kind of linguistic generalizations neural networks are capable of learning from data. We report on an investigation into how \emph{consonant gradation}, a particular morphophonological process which is common in Finnish and other Uralic languages, is encoded in the hidden states of an LSTM encoder-decoder model trained to perform word inflection. Specifically, we train character-based sequence-to-sequence models for inflection of Finnish nouns into the genitive case, an inflection type which commonly triggers consonant gradation. 

\begin{figure}
    \centering
    \includegraphics[width=\columnwidth]{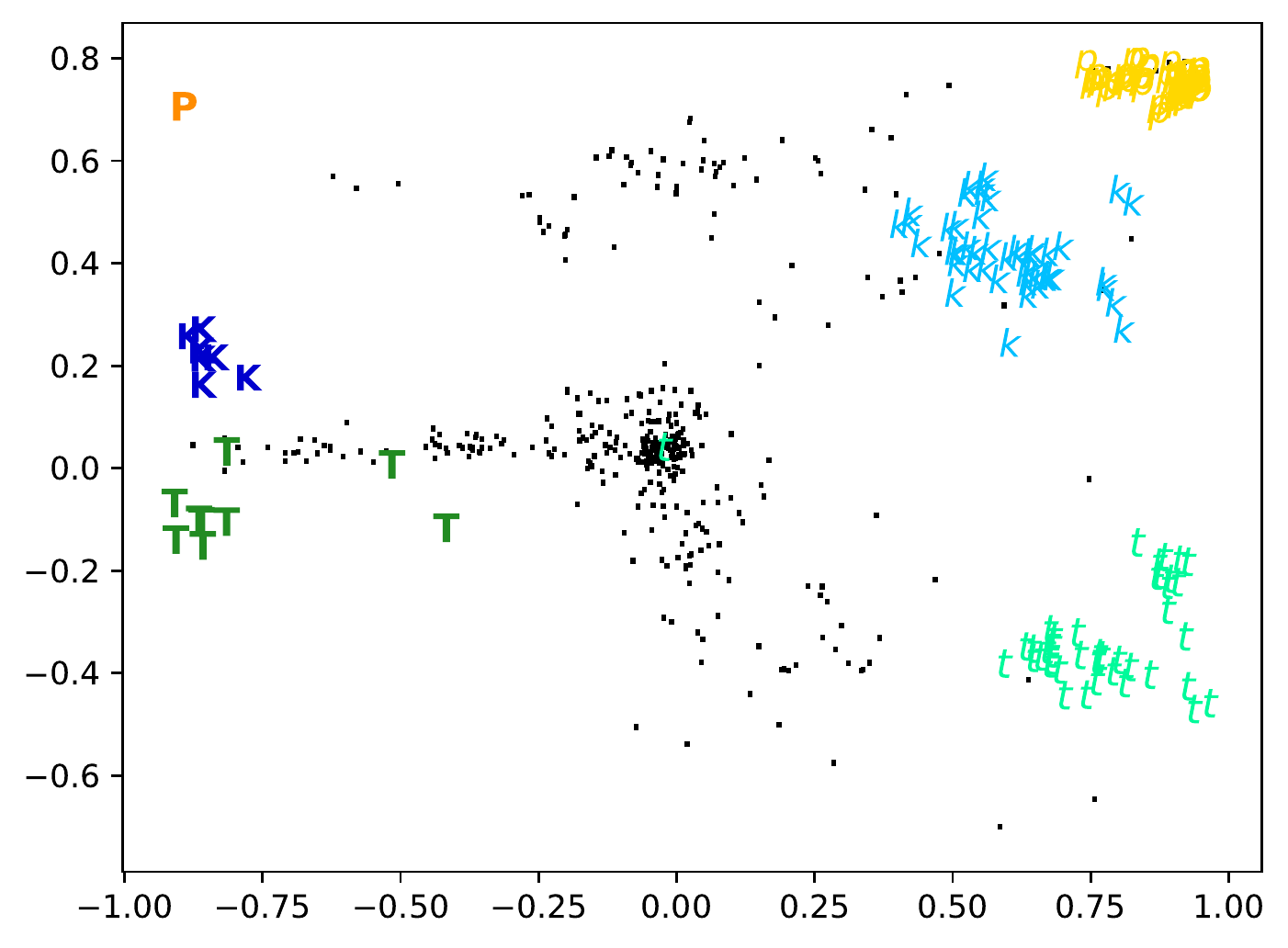}
    \caption{Scatter plot of the activation for two encoder hidden state dimensions which activate strongly during gradation. The letters \textbf{p}, \textbf{t} and \textbf{k} refer to examples which undergo gradation for the respective consonants. Lower-case characters mark direct (weakening) gradation and upper-case characters inverse (strengthening) gradation (see Section \ref{sec:cg} for further details). Black dots  \textbf{·} mark examples which do not undergo gradation.}
    \label{fig:scatterplot}
\end{figure}

Consonant gradation is a phonological process which lenites voiceless stops {\bf p}, {\bf t} and {\bf k} in certain positions (see Section~\ref{sec:cg} for further details). We first demonstrate that inflection networks tend to learn an abstract representation for consonant gradation, where the alternation is triggered by the same dimensions in encoder hidden states regardless of which stop {\bf p}, {\bf t} or {\bf k} undergoes gradation. This echoes the treatment of gradation in linguistic literature \cite[§41]{visk2004iso} Nevertheless, we also find evidence that this behavior is not universal and that networks can sometimes fail to generalize gradation and instead learn to represent gradation using distinct dimensions for each stop {\bf p}, {\bf t} and {\bf k}. 

Our second contribution is to show that networks can learn a general representation encompassing both so-called {\it quantitative gradation} and {\it qualitative gradation} %, where a geminate e.g. {\bf kk} is lenited to the corresponding non-geminate {\bf k}, 
 %where a stop like {\bf t} is lenited to the corresponding voiced stop {\bf d} 
(these are further described in Section \ref{sec:cg}). This presents further evidence that the phonological representations learned by encoder-decoder models can learn to group linguistic generalizations that target different sounds.

As our third contribution, we show evidence of a remarkable property whereby directionality of gradation is encoded as positive or negative hidden state activations: 
Consonant gradation is called direct when the base form of a noun displays the strong grade (such as {\bf kk}) and the genitive form displays the weak grade of a stop (such as {\bf k}). In inverse (or `strengthening') gradation, the opposite alternation occurs. We find hidden state dimensions which encode for the direction of gradation by a positive or negative activation. This behavior is demonstrated in Figure~\ref{fig:scatterplot} where a negative activation of dimension 487 in the encoder hidden state marks inverse gradation of a stop, and positive activation instead marks direct gradation (see Section~\ref{sec:experiments} for further discussion of this phenomenon).

\section{Related Work}

Interpretation of neural representations in recurrent neural models has been an active area of research over a long period of time starting with \newcite{elman1990}.  However, representations in models of phonology have received less attention than many other subfields of NLP.  \newcite{rodd1997recurrent} investigates learning of Turkish vowel harmony by a character-based RNN language model trained on word forms. The paper investigates hidden state activations of RNN models while varying the hidden state dimensionality between 1 and 4. It presents evidence that RNN hidden states can capture Turkish vowel harmony patterns when a sufficient number of hidden dimensions are available.
In a similar vein, \newcite{silfverberg-etal-2018-sound} investigate phoneme representations for Finnish, Spanish and Turkish finding correlations between embedding representations and phonological distinctive features. \newcite{kolachina2019phone} present an investigation of phone embeddings learned using word2vec \cite{mikolov2013distributed} for simulated data showing that phone embeddings capture phonemic and allophonic relationships. They also show that phone embeddings capture co-occurrence restrictions for vowels well, while largely failing to do this for consonants. Our encoder representations, in contrast, are able to capture these co-occurrence restrictions.  

\newcite{beguvs2020modeling} investigates representations learned by a generative adversarial network or GAN \cite{goodfellow2014generative} trained on audio recordings of speech, showing that some of the latent variables of the GAN correspond to phonological features of the speech signal: specifically the presence or absence of the fricative [s] in the output of the network and the amplitude of frication. They show that manipulation of the variables changes these features in a predictable manner. Similarly to our work, \newcite{beguvs2020modeling} also scales state activations and observes the effect on the output of the network. In a related investigation of reduplication, \newcite{beguvs2020identity} train GAN models on speech and identify variables which trigger reduplication in the speech signal. 

Extensive work exists on linguistic probing experiments for neural representations \cite{conneau2018you,conneau2018xnli,clark2019does}. A recent probing paper by \newcite{hennigen2020intrinsic} is more directly related to our work. They present a decomposable probe for finding small sets of hidden states which encode for linguistically relevant information, particularly morphosyntactic information. Our work shares the aim of not only identifying if information is present in a neural system, but also examining how it is represented. However, we additionally perform experiments on manipulating network activations and examine how such manipulations influence the outputs of the network.
% FMT: Commented out for space reasons
%\newcite{hennigen2020intrinsic} use mutual information as a measure for the association of state activations and linguistic features following recent work on probing \cite{voita2020information}. We instead directly measure the difference in state activations at positions where a given linguistic attribute occurs. This direct approach is made possible by the fact that we can identify the location of phonological alternations very reliably in the input which might not be true for linguistic features in general. 
\begin{table}[t]
\centering
\begin{adjustbox}{width=0.85\columnwidth}
\begin{tabular}{llll}
                 \toprule
                  & {\bf Nominative} & {\bf Genitive} & {\bf Gloss} \\
                 \midrule
                  & pa\underline{pp}i & pa\underline{p}in   & `priest'   \\
                  
                  & ken\underline{tt}ä & ken\underline{t}än   & `field'     \\
                  \multirow{2}{*}{\bf Quantitative} 
                  & kiu\underline{kk}u & kiu\underline{k}un   & `anger'     \\
                  & ri\underline{p}e   & ri\underline{pp}een & `remain(s)'     \\
                  & lai\underline{t}e   & lai\underline{tt}een  & `device'     \\
                  & lii\underline{k}e   & lii\underline{kk}een  & `motion'     \\
\midrule
                  & so\underline{p}u & so\underline{v}un  & `agreement'     \\
                  & joh\underline{t}o  & joh\underline{d}on  & `lead'     \\
                  & ai\underline{k}a  & a\underline{j}an   & `time'     \\
                  & ky\underline{k}y & ky\underline{v}yn  & `skill'    \\
                  & olen\underline{t}o & olen\underline{n}on  & `creature' \\
                 {\bf Qualitative} & ken\underline{k}ä & ken\underline{g}än  & `shoe'      \\
                  & sil\underline{t}a & sil\underline{l}an  & `bridge'     \\
                  & rum\underline{p}u & rum\underline{m}un  & `drum'     \\
                  & ran\underline{n}e & ran\underline{t}een  & `wrist'     \\
                  & sal\underline{k}o & salon & `pole'   \\
                  & olu\underline{t}   & oluen  & `beer'     \\
\bottomrule
\end{tabular}
\end{adjustbox}
\caption{\label{tab:gradationexamples}Examples of various kinds of consonant gradation in Finnish.  Adding the genitive suffix {\bf -n} closes the final syllable triggering gradation.}
\end{table}

Our approach was inspired by the now-classic paper on visualization and interpretation of recurrent networks by \newcite{karpathy2015visualizing} in that we also seek individual interpretable dimensions. The work by  \newcite{dalvi2019one} on analyzing individual neurons in networks trained for linguistic tasks (POS tagging as well as semantic and morphological tagging) is more closely related to the present work. They present a general methodology for uncovering neurons which encode linguistic information by training a classifier to predict linguistic features of the input based on the representations generated by the network. They also show that it is possible to manipulate specific neurons to force the network to generate a particular linguistic feature.  
\newcite{meyes2019ablation} investigate the effect of scaling network activations, something they call \emph{ablation} studies. They train classifiers and look at how classification performance varies when zeroing out the activations for particular states or groups of states. 

% FMT: Commented out for space
%Attention in encoder-decoder models  \newcite{bahdanau2014neural} is an active field of study in NLP. Although we do not study attention, works which aim to perturb the attention distribution are relevant in as much as they resemble our approach of scaling hidden state dimensions. \newcite{serrano2019attention} explore the effect of manipulating network attention of classification models and show that attention is at best a very noisy predictor of the important of inputs. %Often the classification decisions of the model will not change when attention is zeroed out.

%TODO: Look up related work in adversarial training.

\section{Consonant Gradation}\label{sec:cg}

Consonant Gradation (CG), common in many Uralic languages, is a set of assimilation and lenition processes, usually targeting the final syllable in a word stem. Historically the trigger for the alternation has been purely phonological, but in Finnish, the process is no longer entirely predictable from the phonological structure \cite{abondolo1998}. The trigger for gradation is usually an affix that closes the final syllable, such as the genitive {\bf -n}, e.g. {\bf ka\underline{tt}o} $\sim$ {\bf ka\underline{t}on} (`roof' sg. nom. $\sim$ sg.\footnote{Underlining marks the position affected by gradation.} gen.). The overall process is divided into {\it quantitative gradation} where, for example, geminate {\bf pp}, {\bf tt}, {\bf kk} alternate with their non-geminate counterparts, {\bf p}, {\bf t}, {\bf k}, and {\it qualitative gradation} where a large variety of lenition and assimilation processes are found. For example, strong grade {\bf k} can alternate with the weakened {\bf j}, {\bf v}, {\bf g}, etc.  %GN: Do we need to define quant., qual. gradation twice?  I prefer the definition here, with just a reference in the intro. 
See Table \ref{tab:gradationexamples} for a summary of these types of gradation processes found in our data set. The lenited or elided forms are commonly called the {\it weak grade} (e.g. {\bf ka\underline{t}on}) and the alternant the {\it strong grade} (e.g. {\bf ka\underline{tt}o}). Sometimes the weak and strong grades appear in the inverse position, i.e. the weak grade appears with open syllables as in {\bf ri{\underline k}e} $\sim$ {\bf ri\underline{kk}een} (`offense' sg. nom. $\sim$ sg. gen.).  

\begin{figure*}
    \centering
    \includegraphics[width=0.7\textwidth]{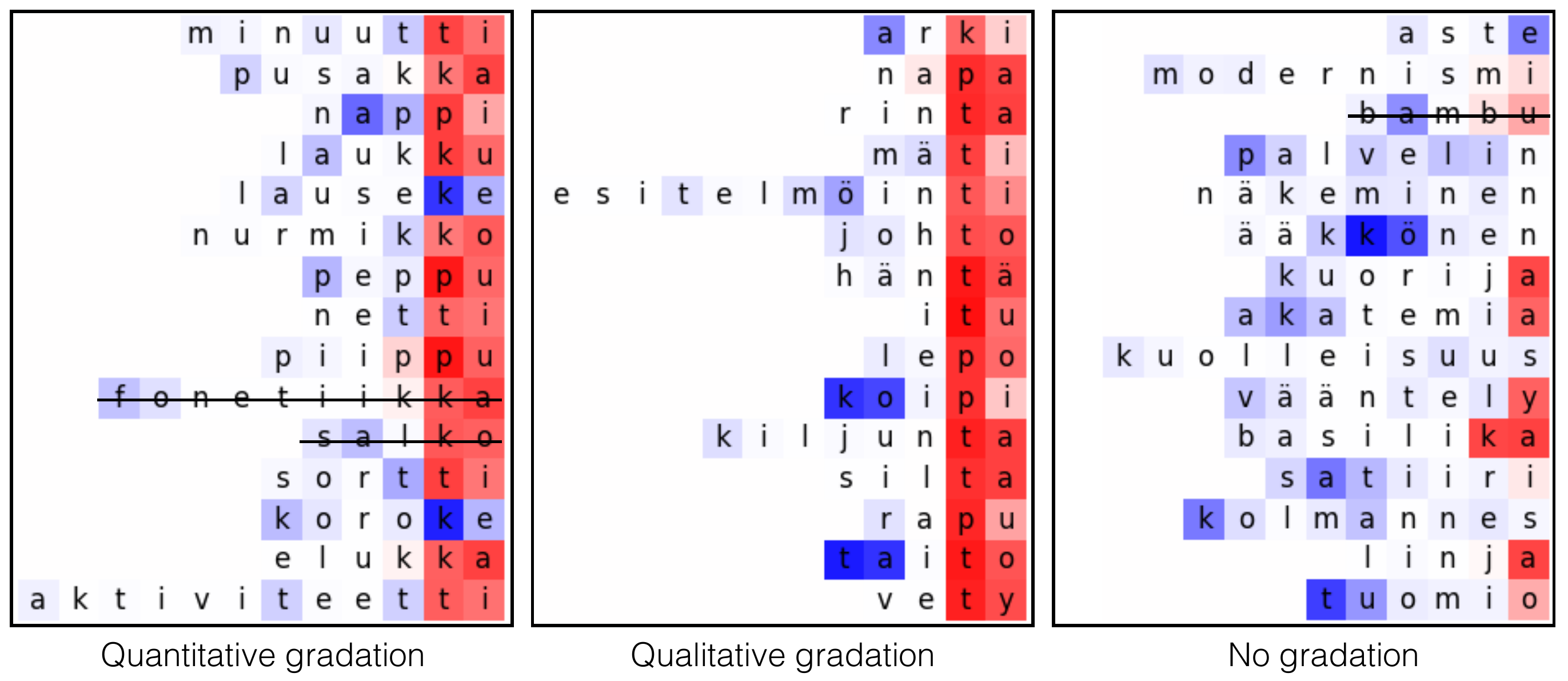}
    \caption{Activation of encoder hidden state dimension 487 in model 3 when (1) quantitative gradation, (2) qualitative gradation and (3) no gradation occurs in the input form. Red squares indicate positive and blue squares negative activation. Forms with strikethrough are inflected incorrectly by the decoder: {\bf *fonetikan} ({\bf fonetiikan}), {\bf *salkon} ({\bf salon}) and {\bf *bammun} ({\bf bambun}).}
    \label{fig:heatmaps}
\end{figure*}

While quantitative gradation remains productive in the language, many stems from more recent loanwords in particular, do not tend to alternate qualitatively; for example {\bf au{\underline t}o} $\sim$ {\bf au\underline{t}on}, ${}^*${\bf au\underline{d}on} (`car' sg. nom. $\sim$ sg. gen.). Speakers must therefore know the lexical status of each stem to inflect it correctly.  Our data set includes both gradating and non-gradating lexemes.

The advantages of studying Finnish consonant gradation in this context is that the set of sound changes is very diverse, but that the trigger for all of them is the same. Also, the Finnish writing system is very phonemic and surface-oriented and therefore no conversion to an IPA representation is necessary to reveal the sound changes that occur as a result of gradation.

Of particular interest to us is that there are many similar-looking alternations in Finnish that are not a result of consonant gradation, but paradigmatic variation. For example, {\bf varis} (`crow' sg. nom.) is inflected {\bf vari\underline{k}sen} in the sg. gen. form.  Note the similarity of this alternation to the actual CG case of {\bf liike} (`motion' sg. nom.) $\sim$ {\bf lii\underline{kk}een} (sg. gen.) which also involves a $\emptyset$ $\sim$ k alternation. It is therefore of some interest to observe whether neural inflection models encode the two cases differently in some respect.

In total we count 17 different types of lenition or fortition falling under the rubric of consonant gradation in our data set; an example of each type is shown in Table \ref{tab:gradationexamples}. 

\section{Methods}
% FMT: Not sure what goes here...

% In order to investigate the representations discovered for the phenomenon of consonant gradation we conducted an experiment in predicting the genitive form of a noun given the nominative form. For example given the word \textbf{lumikko} `weasel' generate the genitive form \textbf{lumikon} `weasel gen. sg.'. In Finnish if the word undergoes gradation it will undergo it in this transformation.

This section presents our nominative $\rightarrow$ genitive inflection models and our approach to finding encoder hidden state dimensions which are associated with consonant gradation. 

\subsection{Inflection Models}

As our inflection model, we use the well-known attentional BiLSTM encoder-decoder model which was presented by \newcite{bahdanau2014neural} and first applied to inflection by \newcite{kann2016med}. This neural model transduces a nominative input form which is represented as a sequence of characters $x_{[1:T]}$ of length $T$ into a genitive output form $y_{[1:S]}$ of length $S$. 

The encoder network in an attentional model generates one hidden state vector ${\bf h}_t = {\bf f}_t \oplus {\bf b}_t \in \mathbb{R}^{2n}$ for every position in the input sequence. Due to the bidirectionality of the encoder, the hidden state vector is a concatenation of a forward state ${\bf f}_t \in \mathbb{R}^n$ and a backward state ${\bf b}_t \in \mathbb{R}^n$. We refer to the vectors ${\bf f}_t$ as hidden states and the elements in the vectors ${\bf f}_t[d]$ as {\it activations}. Here $d \in \{1, 2, 3, ..., 2n\}$ is called a {\it dimension}.

% TODO: how we find the states

\subsection{Finding Dimensions Associated with Gradation}
\label{sec:finding-act}

Our aim is to investigate encoder hidden state dimensions $d$ which are associated with gradation. To this end, we extract the encoder hidden state activations ${\bf h}_1[d]$, ..., ${\bf h}_T[d]$ for each example $(x_{[1:T]},y_{[1:S]})$ in our development set and dimension $d\in \{1,2,3, ..., 2n\}$. In order to find dimensions which activate strongly at positions where gradation occurs, we compare the mean activation of each dimension for forms which undergo gradation and forms which do not.  

Let $a_X(d)$, as defined by Equation~(\ref{eq:mean-act}) below, be the mean activation for dimension $d$ in a set of encoder hidden states $X$. For each dimension $d$, we extract the mean activation $a_G(d)$, where $G$ is the set of encoder hidden states at positions where gradation occurs. As explained in Section \ref{sec:cg}, gradation applies to the final stop in word forms which undergo gradation. Usually, this would refer to position $T-1$ in a string of length $T$ as in {\bf tu\underline{p}a} `cottage sg. nom.', where {\bf p} undergoes gradation, but can also happen at position $T-2$ as in the form {\bf ra\underline{t}as} `wheel sg. nom.', where {\bf t} undergoes gradation.

\begin{equation}
    a_X(d) = \frac{\sum_{h\in X} {\bf h}[d]}{|X|} 
    \label{eq:mean-act}
\end{equation}

The mean activation $a_G(d)$ is compared to the activation $a_N(d)$ of dimension $d$ at the penultimate position $T-1$ in base forms of length $T$ which do not undergo gradation. In order to specifically capture dimensions which encode for gradation as opposed to simply encoding for consonants, we limit this examination to base forms like {\bf kana} `chicken sg. nom.' and {\bf auto} `car sg. nom.', where the penultimate character is a consonant. %---which may or may not undergo gradation. 
We retrieve the top-$N$ dimensions $d$ where the difference in mean activation $|a_N(d) - a_G(d)|$ is maximized and consider these candidate dimensions for gradation.

\section{Data}

Our dataset was produced by taking the most frequent 5,000 lexemes tagged as singular nominative nouns from the Turku Dependency Treebank \cite{haverinen:13} and generating the singular genitive forms using the OmorFi finite-state morphological transducer \cite{pirinen2015development}. We excluded compound nouns (e.g. \textbf{ammattikorkeakoulututkinnoista} `from the professional high-school examinations') and words marked as nouns which contained punctuation or numerals (e.g. \textbf{G8-neuvottelut} `G8 negotiations', \textbf{2000-luvulla} `in the 2000s', \textbf{°C:ssa} `in °C' etc.). Loan words were included, both unadapted such as \textbf{workshop} and \textbf{bungalow} and partially or fully adapted such as \textbf{brosyyri} `brochure' and \textbf{samppanja} `champagne'. This gave a total of 4,797 nominative--genitive pairs. We randomly ordered them and then split these into disjoint sets: 90\% for training (4,317 pairs) and 10\% for validation.

We then took the validation set (479 pairs) and annotated them for: gradation (yes, no), type of gradation (qualitative, quantitative), consonant ({\bf p}, {\bf t}, {\bf k}) and direction (direct, inverse). This gave a total of 84 examples of nouns exhibiting consonant gradation. 

This set was heavily skewed towards {\bf t} gradation (54 out of 84 examples).\footnote{This follows character-level frequency patterns in Finnish, e.g. in the treebank {\bf t} appears 122,821 times, {\bf k} appears 64,513 times and {\bf p} appears 23,130 times.} So we randomly sampled another 84 words from the frequency list, which were not found in the training data or in the existing validation set and which contained {\bf p} and {\bf k}, and annotated them and added them to the validation set.  Statistics on the composition of the hand-annotated dataset can be found in Table~\ref{tab:validset} and the full data is freely available on GitHub.\footnote{\url{https://github.com/mpsilfve/gradation}}

\begin{table}[]
    \centering
    \begin{adjustbox}{width=0.65\columnwidth}
    \begin{tabular}{lrrrr}
    \toprule
            &    \multirow{2}{*}{\textbf{Ungrad.}} & \multicolumn{3}{c}{\textbf{Grad.}} \\
              &   &          Qual. & Quant.  & Total \\
              \midrule
          {\bf p}  & 35 & 14 & 40  & 54 \\ %inv 1
         {\bf t}  & 159 & 24 & 30 & 54 \\ %inv 0
         {\bf k}  & 128 & 4 & 50  & 54 \\ %inv 6

         \midrule
         \textbf{Total:} & 322 & 42 & 120  & 162 \\
         \bottomrule
    \end{tabular}
    \end{adjustbox}
    \caption{Composition of the manually annotated validation set. The \textbf{Ungrad.} column refers to forms that do not exhibit gradation in the genitive, \textbf{Grad.} those forms that exhibit gradation and \textbf{Qual.} is qualitative and \textbf{Quant.} is quantitative gradation.}
    \label{tab:validset}
\end{table}

\section{Experiments and Results}
\label{sec:experiments}

We investigate representation of consonant gradation in encoder hidden states in the following way: As explained in Section \ref{sec:finding-act}, we identify individual dimensions in encoder hidden states which activate strongly during gradation regardless of the identity of the consonant undergoing gradation. We then investigate the association of these states using two experiments: we (1) perform significance tests on a held-out dataset to determine if the states activate significantly more strongly when gradation occurs, and (2)  scale the state activations and observe the effect on the output of the network. %Our approach identifies individual states which are strongly associated to gradation for all stops {\bf k}, {\bf p} and {\bf t}. Moreover, we notice that we can prevent gradation from occurring when the model is inflecting a word by scaling the gradation-controlling neurons appropriately.

\subsection{Training Details}

\begin{figure}
\begin{center}
    \includegraphics[width=\columnwidth]{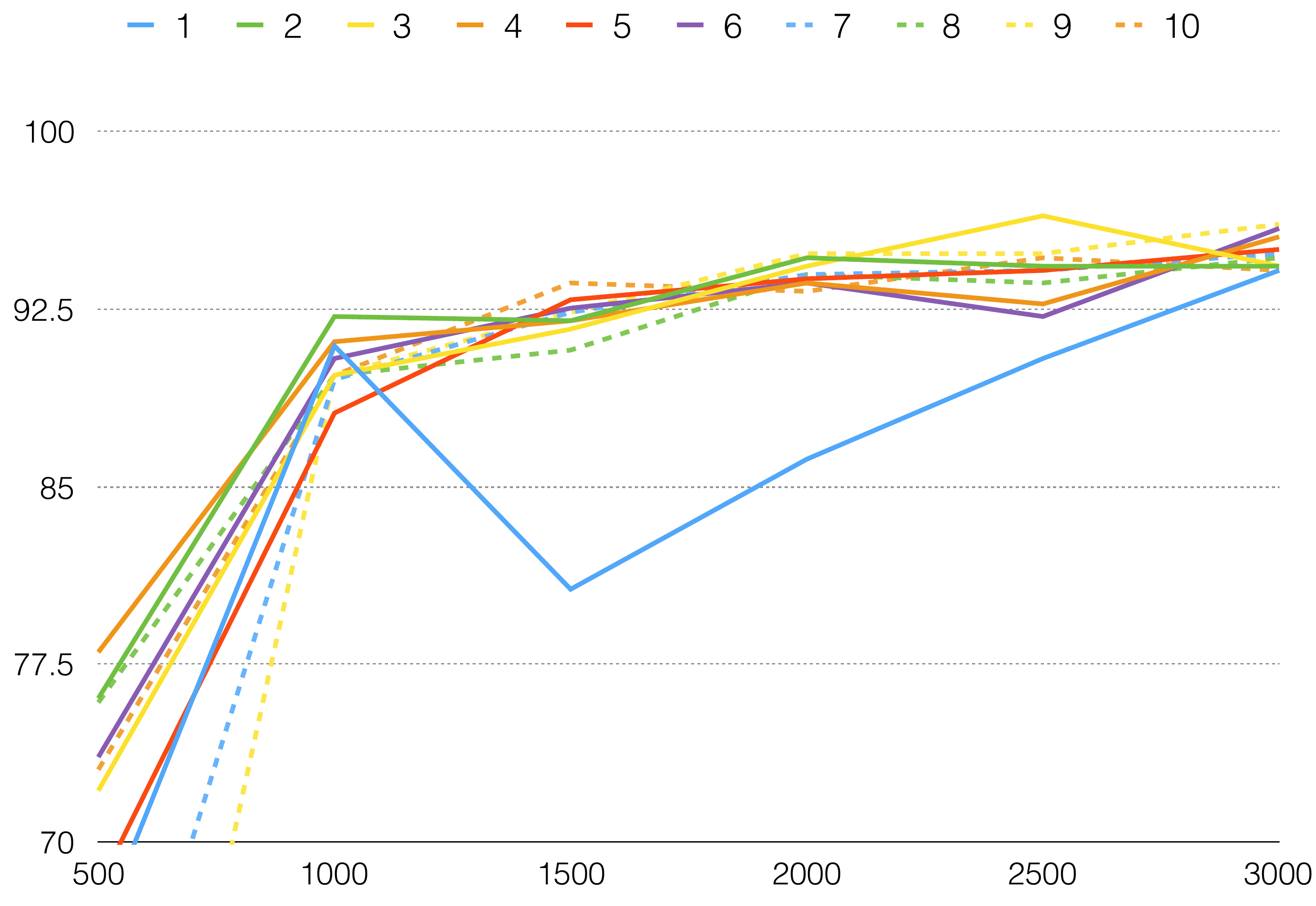}
    \end{center}
    \caption{Percent inflection accuracy on the development set for steps 500-3,000 during the training of our ten inflection models. The figure demonstrates that changes in accuracy on the development set are modest after step 2,000.}
    \label{fig:convergence}
\end{figure}

We train ten encoder-decoder models with different random initializations for inflection using the OpenNMT toolkit \cite{klein-etal-2018-opennmt}. We use a 2-layer BiLSTM encoder with hidden dimension 250. Due to the bidirectionality of the encoder this results in 500-dimensional hidden states (consisting of a forward and backward hidden state). Our model uses 500-dimensional character embeddings both in the encoder and decoder and we use an attentional decoder with 250-dimensional hidden states. The model is trained for a total of 3,000 steps using stochastic gradient descent and a batch size of 64. See Figure \ref{fig:convergence} for a plot of the development accuracy during the training process. As can be seen, changes in development accuracy are modes after training step 2,000.

We report inflection accuracy for our ten inflection models measured on held-out data in Table \ref{tab:infl-acc}. The accuracy is reported separately for forms undergoing gradation and forms not undergoing gradation. In addition, we report an overall accuracy for all forms. We can see that the mean performance is close 95\% for all forms and performance tends to be higher on forms undergoing gradation than other forms.

\begin{table}[!t]
    \begin{center}
    \begin{adjustbox}{width=0.85\columnwidth}
    \begin{tabular}{lrrrr}
    \toprule
        \textbf{Model} & \textbf{\# States} & \textbf{+Grad.} & \textbf{-Grad.} & \textbf{Overall}  \\
        \midrule
       1  & 2 & 94.4 & 94.0 & 94.1 \\
       2  & 2  & 95.1       & 94.0     & 94.3    \\
       3  & 1  & 94.4       &  94.3    & 94.3   \\
       4  & 3  & 96.9       & 95.0     & 95.6 \\
       5  & 3  & 95.1       &  95.0    & 95.0   \\
       6  & 3  & 97.5       & 95.3     &95.9 \\
       7  & 1  & 95.7       & 94.5     &94.9 \\
       8  &  0 & 95.7       & 94.3     &94.7\\
       9  &  0 & 96.9       &  95.8    &96.1 \\
       10  & 2  &  96.9     & 93.0     &94.1 \\
       \bottomrule
    \end{tabular}
    \end{adjustbox}
    \end{center}
    \caption{Percent inflection accuracy for 10 NOM to GEN models trained using different random seeds. The column \# States refers to the number of states found in Table~\ref{tab:significance} that have significant activations for all gradation types.}
    \label{tab:infl-acc}
\end{table}

\begin{table*}[]
\centering
\begin{adjustbox}{width=0.45\textwidth}
\begin{tabular}{lccccc}
\toprule
\multicolumn{6}{c}{{\sc Model 1}}\\
\toprule
\multirow{2}{*}{\textbf{Gradation}}       & \multicolumn{5}{c}{\textbf{Dimension}}\\
 & \colorbox{gray!30}{\bf 458} & \colorbox{gray!30}{\bf 252} & 395 & 489 & 396\\
\midrule
K & {\bf0.247} &{\bf0.162} &0.048 &{\bf0.156} &{\bf0.649} \\
P & {\bf0.291} &{\bf0.350} &{\bf0.390} &{\bf0.283} &0.056 \\
T & {\bf0.175} &{\bf0.062} &0.022 &0.043 &{\bf0.193} \\
Qual. & {\bf0.259} &{\bf0.176} &{\bf0.200} &{\bf0.136} &0.069 \\
Quant. & {\bf0.252} &{\bf0.218} &{\bf0.199} &{\bf0.194} &{\bf0.245} \\
\bottomrule
\end{tabular}\end{adjustbox}~~~\begin{adjustbox}{width=0.45\textwidth}
\begin{tabular}{lccccc}
\toprule
\multicolumn{6}{c}{{\sc Model 2}}\\
\toprule
\multirow{2}{*}{\textbf{Gradation}}       & \multicolumn{5}{c}{\textbf{Dimension}}\\
 & 255 & 127 & 453 & \colorbox{gray!30}{\bf 399} & \colorbox{gray!30}{\bf 469}\\
\midrule
K & 0.080 &{\bf0.285} &{\bf0.395} &{\bf0.319} &{\bf0.281} \\
P & {\bf0.557} &{\bf0.354} &{\bf0.371} &{\bf0.260} &{\bf0.228} \\
T & {\bf0.168} &0.158 &{\bf0.122} &{\bf0.233} &{\bf0.277} \\
Qual. & {\bf0.330} &0.047 &0.132 &{\bf0.242} &{\bf0.224} \\
Quant. & {\bf0.314} &{\bf0.368} &{\bf0.335} &{\bf0.277} &{\bf0.252} \\
\bottomrule
\end{tabular}
\end{adjustbox}

\begin{adjustbox}{width=0.45\textwidth}
\begin{tabular}{lccccc}
\multicolumn{6}{c}{{\sc Model 3}}\\
\toprule
\multirow{2}{*}{\textbf{Gradation}}       & \multicolumn{5}{c}{\textbf{Dimension}}\\
 & 487 & 207 & \colorbox{gray!30}{\bf 203} & 484 & 221\\
\midrule
K & 0.205 &{\bf0.412} &{\bf0.393} &{\bf0.263} &{\bf0.276} \\
P & {\bf0.717} &{\bf0.471} &{\bf0.248} &{\bf0.622} &{\bf0.366} \\
T & {\bf0.567} &0.080 &{\bf0.324} &{\bf0.475} &0.127 \\
Qual. & {\bf0.684} &{\bf0.254} &{\bf0.268} &0.118 &{\bf0.249} \\
Quant. & {\bf0.440} &{\bf0.423} &{\bf0.323} &{\bf0.358} &{\bf0.289} \\
\bottomrule
\end{tabular}
\end{adjustbox}~~~\begin{adjustbox}{width=0.45\textwidth}
\begin{tabular}{lccccc}
\multicolumn{6}{c}{{\sc Model 4}}\\
\toprule
\multirow{2}{*}{\textbf{Gradation}}       & \multicolumn{5}{c}{\textbf{Dimension}}\\
 & \colorbox{gray!30}{\bf349} & \colorbox{gray!30}{\bf163} & 124 & \colorbox{gray!30}{\bf416} & 254\\
\midrule
K & {\bf0.229} &{\bf0.498} &{\bf0.540} &{\bf0.078} &{\bf0.364} \\
P & {\bf0.260} &{\bf0.081} &{\bf0.115} &{\bf0.414} &0.099 \\
T & {\bf0.386} &{\bf0.193} &{\bf0.082} &{\bf0.070} &0.139 \\
Qual. & {\bf0.222} &{\bf0.122} &0.006 &{\bf0.294} &0.068 \\
Quant. & {\bf0.270} &{\bf0.290} &{\bf0.297} &{\bf0.214} &{\bf0.273} \\
\bottomrule
\end{tabular}
\end{adjustbox}

\begin{adjustbox}{width=0.45\textwidth}
\begin{tabular}{lccccc}
\multicolumn{6}{c}{{\sc Model 5}}\\
\toprule
\multirow{2}{*}{\textbf{Gradation}}       & \multicolumn{5}{c}{\textbf{Dimension}}\\
 & \colorbox{gray!30}{\bf345} & \colorbox{gray!30}{\bf109} & 408 & 390 & \colorbox{gray!30}{\bf191}\\
\midrule
K & {\bf0.272} &{\bf0.206} &{\bf0.775} &0.041 &{\bf0.225} \\
P & {\bf0.466} &{\bf0.381} &{\bf0.105} &{\bf0.498} &{\bf0.183} \\
T & {\bf0.498} &{\bf0.186} &0.057 &{\bf0.161} &{\bf0.257} \\
Qual. & {\bf0.397} &{\bf0.241} &0.068 &{\bf0.206} &{\bf0.162} \\
Quant. & {\bf0.395} &{\bf0.296} &{\bf0.298} &{\bf0.224} &{\bf0.222} \\
\bottomrule
\end{tabular}
\end{adjustbox}~~~\begin{adjustbox}{width=0.45\textwidth}
\begin{tabular}{lccccc}
\multicolumn{6}{c}{{\sc Model 6}}\\
\toprule
\multirow{2}{*}{\textbf{Gradation}}       & \multicolumn{5}{c}{\textbf{Dimension}}\\
 & \colorbox{gray!30}{\bf480} & \colorbox{gray!30}{\bf22} & 418 & \colorbox{gray!30}{\bf48} & 320\\
\midrule
K & {\bf0.303} &{\bf0.240} &{\bf0.257} &{\bf0.191} &{\bf0.651} \\
P & {\bf0.394} &{\bf0.353} &{\bf0.346} &{\bf0.297} &0.045 \\
T & {\bf0.593} &{\bf0.073} &0.012 &{\bf0.198} &0.055 \\
Qual. & {\bf0.320} &{\bf0.140} &{\bf0.180} &{\bf0.162} &0.018 \\
Quant. & {\bf0.388} &{\bf0.294} &{\bf0.275} &{\bf0.262} &{\bf0.267} \\
\bottomrule
\end{tabular}
\end{adjustbox}

\begin{adjustbox}{width=0.45\textwidth}
\begin{tabular}{lccccc}
\multicolumn{6}{c}{{\sc Model 7}}\\
\toprule
\multirow{2}{*}{\textbf{Gradation}}       & \multicolumn{5}{c}{\textbf{Dimension}}\\
 & 142 & 108 & 230 & 207 & \colorbox{gray!30}{\bf405}\\
\midrule
K & {\bf0.212} &{\bf0.223} &{\bf0.138} &0.068 &{\bf0.309} \\
P & {\bf0.319} &{\bf0.452} &{\bf0.421} &{\bf0.438} &{\bf0.169} \\
T & 0.172 &{\bf0.227} &0.032 &0.028 &{\bf0.122} \\
Qual. & 0.180 &0.099 &0.148 &0.170 &{\bf0.166} \\
Quant. & {\bf0.268} &{\bf0.282} &{\bf0.278} &{\bf0.232} &{\bf0.226} \\
\bottomrule
\end{tabular}
\end{adjustbox}~~~\begin{adjustbox}{width=0.45\textwidth}
\begin{tabular}{lccccc}
\multicolumn{6}{c}{{\sc Model 8}}\\
\toprule
\multirow{2}{*}{\textbf{Gradation}}       & \multicolumn{5}{c}{\textbf{Dimension}}\\
 & 441 & 234 & 238 & 283 & 125\\
\midrule
K & 0.033 &{\bf0.272} &{\bf0.191} &0.075 &0.013 \\
P & {\bf0.463} &{\bf0.243} &{\bf0.240} &{\bf0.352} &{\bf0.260} \\
T & 0.038 &0.085 &0.025 &0.087 &{\bf0.241} \\
Qual. & {\bf0.238} &0.080 &{\bf0.154} &0.149 &{\bf0.192} \\
Quant. & {\bf0.238} &{\bf0.227} &{\bf0.193} &{\bf0.185} &{\bf0.164} \\
\bottomrule
\end{tabular}
\end{adjustbox}

\begin{adjustbox}{width=0.45\textwidth}
\begin{tabular}{lccccc}
\multicolumn{6}{c}{{\sc Model 9}}\\
\toprule
\multirow{2}{*}{\textbf{Gradation}}       & \multicolumn{5}{c}{\textbf{Dimension}}\\
 & 446 & 459 & 438 & 380 & 351\\
\midrule
K & 0.184 &{\bf0.375} &{\bf0.487} &{\bf0.157} &{\bf0.109} \\
P & {\bf0.416} &{\bf0.293} &{\bf0.142} &{\bf0.283} &{\bf0.379} \\
T & {\bf0.208} &0.030 &{\bf0.075} &0.124 &{\bf0.135} \\
Qual. & {\bf0.258} &{\bf0.137} &0.044 &{\bf0.242} &0.117 \\
Quant. & {\bf0.293} &{\bf0.314} &{\bf0.282} &{\bf0.196} &{\bf0.220} \\
\bottomrule
\end{tabular}
\end{adjustbox}~~~\begin{adjustbox}{width=0.45\textwidth}
\begin{tabular}{lccccc}
\multicolumn{6}{c}{{\sc Model 10}}\\
\toprule
\multirow{2}{*}{\textbf{Gradation}}       & \multicolumn{5}{c}{\textbf{Dimension}}\\
 & \colorbox{gray!30}{\bf56} & 270 & 284 & \colorbox{gray!30}{\bf367} & 306\\
\midrule
K & {\bf0.563} &0.172 &0.108 &{\bf0.357} &{\bf0.438} \\
P & {\bf0.153} &{\bf0.326} &{\bf0.429} &{\bf0.145} &0.132 \\
T & {\bf0.194} &{\bf0.177} &0.064 &{\bf0.076} &0.075 \\
Qual. & {\bf0.173} &{\bf0.240} &{\bf0.215} &{\bf0.120} &0.158 \\
Quant. & {\bf0.348} &{\bf0.232} &{\bf0.218} &{\bf0.235} &{\bf0.305} \\
\bottomrule
\end{tabular}
\end{adjustbox}

    \caption{Mean differences in activation strength for dimension $d$ where we first find the top-5 states associated with gradation using 50\% of the development data and then perform significance tests using the remaining 50\% of the development data. We present results for 10 different random initializations of model parameters. We compare activation when {\bf k}, {\bf p} or {\bf t} gradation occurs to activation at {\bf -CV} word endings where gradation does not occur. We also report results for qualitative and quantitative gradation irrespective of the consonant undergoing gradation. Statistically significant differences in activation strength at the 99.5\% significance level are shown in bold face. Dimensions with significantly stronger association for all stops as well as qualitative and quantitative gradation are marked using a \colorbox{gray!30}{{\bf gray box}}.}
    \label{tab:significance}
\end{table*}

%FMT: I rotated this table, i think it makes it a bit easier to read and compare numbers this way, but feel free to revert/edit.

%\begin{table}[]
%    \centering
%    \begin{adjustbox}{width=\columnwidth}
%    \begin{tabular}{cccccccccc}
%    \multicolumn{10}{c}{{\sc All forms}}\\
%        1 & 2 & 3 & 4 & 5 & 6 & 7 & 8 & 9 & 10\\
%        \midrule
%        94.1 & 94.3 & 94.3 & 95.6 & 95.0 & 95.9 & %94.9 & 94.7 & 96.1 & 94.1\\ 
%        \\
%      \multicolumn{10}{c}{{\sc Forms undergoing gradation}}\\
%        1 & 2 & 3 & 4 & 5 & 6 & 7 & 8 & 9 & 10\\
%        \midrule
%        94.4 & 95.1 & 94.4 & 96.9 & 95.1 & 97.5 & 95.7 & 95.7 & 96.9 & 96.9\\ 
%        \\
  %  \multicolumn{10}{c}{{\sc Forms which do not undergo gradation}}\\
%        1 & 2 & 3 & 4 & 5 & 6 & 7 & 8 & 9 & 10\\
 %       \midrule
%        94.0 & 94.0 & 94.3 & 95.0 & 95.0 & 95.3 & 94.5 & 94.3 & 95.8 & 93.0\\  
%      \multicolumn{10}{c}{{\sc `g' states}}\\   
%              1 & 2 & 3 & 4 & 5 & 6 & 7 & 8 & 9 & 10\\
%        \midrule
%              2&2&1&3&3&3&1&0&0&2 \\
    %\end{tabular}
    %\end{adjustbox}
    %\caption{Inflection accuracy for 10 NOM to GEN models trained using different random seeds.}
%    \label{tab:infl-acc}
%\end{table}

% TODO: Train models with smaller hidden state size 

\subsection{Investigation of State Activations}

 We randomly split our development set into two disjoint parts of equal size. The first part of the development set we use to discover the top-5 encoder hidden state dimensions which are strongly associated with gradation (as described in Section \ref{sec:finding-act}). The rest of the development set is used for significance testing. We perform a two-sided t-test to check if the mean activations of our top-5 dimensions differ significantly (at the 99.5\% significance level) between positions which undergo gradation and positions which do not undergo gradation. As explained in Section \ref{sec:finding-act}, we limit this examination to nominative forms where the penultimate character is a consonant to better zone in on gradation.

Table \ref{tab:significance} shows the results separately for {\bf p}, {\bf t} and {\bf k} gradation. The table also shows results for qualitative and quantitative gradation. We can see that eight of the ten models contain at least one dimension where activation is significantly stronger for all stops  {\bf p}, {\bf t} and {\bf k} undergoing gradation than other stem-final consonants indicating that these states are associated with gradation in general rather than gradation of one of the individual consonants {\bf p}, {\bf t}, or {\bf k}. We note that these dimensions also typically activate both for qualitative and quantitative gradation indicating that the network has learned an abstraction for both types of gradation.

\subsection{Scaling State Activations}
\label{sec:scaling}

As a direct test of the effect of hidden state dimensions on gradation, we scale the activations of dimensions which are strongly associated with gradation. Our hypothesis is that negatively scaling these dimensions will prevent forms from undergoing gradation. 

We experiment on a dataset consisting of all development examples which undergo gradation. For each nominative input form such as \textbf{luu\underline{kk}u}, we identify the correct gold standard genitive form \textbf{luu\underline{k}un} (where {\bf kk} $\rightarrow$ {\bf k} alternation has applied) and an alternate output form \textbf{*luu\underline{kk}un} which is correct apart from the fact that the form has not undergone gradation. We then compute (1) the number of gold standard forms, (2) the number of alternate forms, and (3) the number of nonce forms generated by our models. Nonce forms here refer to erroneous outputs like {\bf *luukuukuukkun} which do not belong in category (2).

We scale the hidden state activations at positions where gradation occurs, that is at the final stop in the nominative form, before feeding the encoder hidden states into the decoder. For each input form, we scale the top-$N$ encoder hidden states which are associated with gradation according to the mapping $a \mapsto x\cdot a$ where $x$ varies between 1 and -25. The number of states which are scaled (that is $N$) is tuned for maximal effect on the number of alternate forms which are generated.  

\begin{figure}[!t]
    \centering
    \includegraphics[width=0.9\columnwidth]{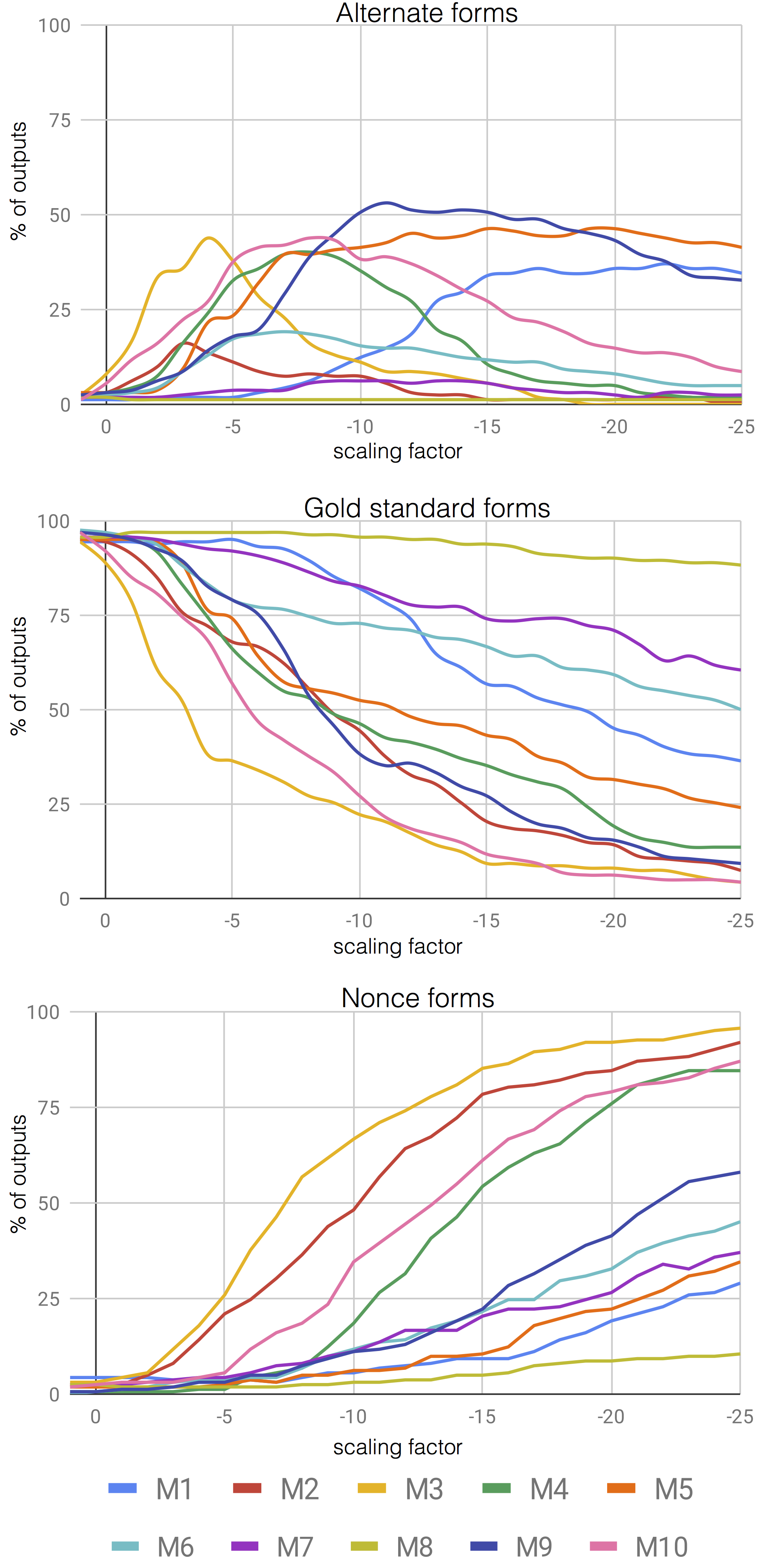}
    \caption{The amount (in \%) of gold standard outputs, alternate outputs not displaying gradation and nonce forms which are generated when scaling top encoder hidden dimensions associated with gradation.}
    \label{fig:scaling_acc}
\end{figure}

Figure \ref{fig:scaling_acc} shows the results for the scaling experiment when tuning $N$.\footnote{For completeness, all results for scaling the top-$N$ states where $N$ varies between 1 and 5 are shown in Appendix \ref{sec:appendix}.}
 %The "Tuned NG" graph shows the number of alternate output forms, where gradation has not been triggered. 
 The first graph shows that for most models the number of alternate forms first increases when the scaling factor $x$ approaches $-25$, and then gradually decreases. As the number of alternate forms increases, the number of gold standard forms undergoing gradation naturally decreases as demonstrated by the second graph. We also see an increase in the number of nonce forms which do not belong to either category. This is to be expected as scaling represents a deviation from learned model weights which disturbs the network.

% \begin{figure}
%     \centering
%     \includegraphics[width=\columnwidth]{state_scaling_accuracy}
%     \caption{Effect of scaling encoder hidden state activations for the top-20 encoder hidden states which are associated with gradation. The green bar indicates the number of examples where the system generated the correct gold standard output form, where gradation occurs. The blue bar instead indicates that the system generated a plausible form which might be correct apart from the fact that gradation did not occur. TODO: Add labels to axes.}
%     \label{fig:scaling-acc}
% \end{figure}

\begin{figure}
    \centering
    \includegraphics[width=\columnwidth]{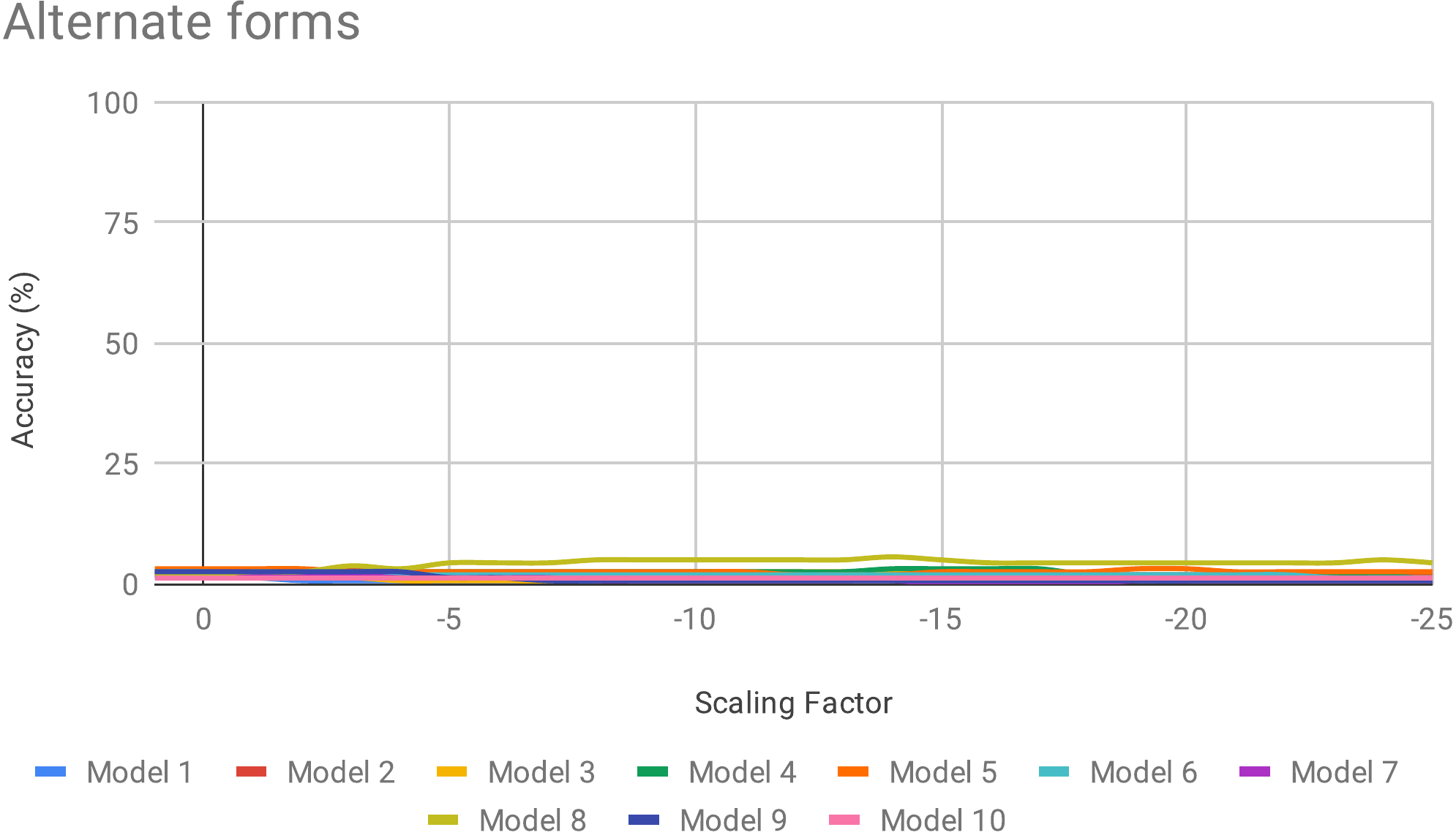}
    \caption{The amount (in \%) of alternate outputs not displaying gradation when five randomly sampled encoder hidden dimensions are scaled.}
    \label{fig:random-dims}
\end{figure}

The effect of scaling varies between models: When scaling activations for Model 9, over half of the output forms do not undergo gradation. In contrast, for Model 7, the best scaling factor only produces around 7\% of non-gradating output forms. Crucially, however, we do see an effect for nearly all models (apart from model 8). Contrast this with Figure \ref{fig:random-dims} which shows results when scaling a set of five random states instead of states which are associated with gradation, showing that scaling of randomly sampled states has very small if any effect on the number of alternate forms produced by the models. Based on the graphs in Figure \ref{fig:scaling_acc}, scaling has very limited effect on Model 8. Even when scaling by $a=-25$, there is only a small decrease in the number of gold standard forms and a corresponding small increase in nonce forms. This might be evidence of a more redundant representation of information in Model 8, whereby scaling a few
states will not strongly perturb the network.

\section{Discussion and Conclusions}

Figure \ref{fig:heatmaps} shows the activation for a hidden state dimension which is strongly associated with gradation: dimension 487 in model 3. This dimension displays positive activation for consonants undergoing direct gradation as in \textbf{lau\underline{kk}u} `bag sg. nom.' $\sim$ \textbf{lau\underline{k}un} `bag sg. gen.'. Remarkably, the state displays negative activation for consonants undergoing inverse gradation as in the example \textbf{lause\underline{k}e} `phrase' where {\bf k} is strengthened into a geminate {\bf kk} resulting in the genitive form \textbf{lause\underline{kk}een} `phrase-{\sc gen}'. This effect can be seen both in forms where quantitative and qualitative gradation occurs. However, as the example \textbf{basilika} `basil' in the third heat map demonstrates, dimension 487 can also activate strongly when no gradation occurs.\footnote{The form {\bf basilika} is a loan word and would probably undergo gradation if it were a native Finnish word. It is noteworthy, however, that regardless of the strong activation of state 487, our model still correctly inflects {\bf basilika} into {\bf basilikan} instead of applying gradation, which would give a form like {\bf *basilijan} or {\bf *basilian}.} This prompted us to investigate hidden state activations more directly using the scaling experiments described in Section~\ref{sec:scaling}.

Figure~\ref{fig:scatterplot} shows a scatter plot of two encoder hidden state dimensions (487 and 484 in model 3) which activate strongly during gradation. Each point in the plot corresponds to one example in our development dataset.  Clearly, examples which do not undergo gradation cluster around \((0,0)\).\footnote{The single \textbf{t} at \((0,0)\) represents the pair \textbf{olu\underline{t}} $\sim$ \textbf{oluen}, where \textbf{t} $\rightarrow$ $\emptyset$. This is an extremely infrequent gradation type.} In contrast, gradation for \textbf{k} and \textbf{p} lead to a positive activation for state 484, whereas \textbf{t}-gradation gives a negative activation. Moreover, direct gradation results in a positive activation for state 487 and inverse gradation gives a negative activation. Examples which do not undergo gradation can also have high values for 484 ($> 0.4$). Many of these examples end in \textbf{-jV}, \textbf{-vV} or \textbf{-mV} which could actually be examples where inverse gradation occurs but it happens not to be the case for these particular ones. Examples where the activation for 484 is low ($< -0.5$) span a small number of forms ending \textbf{-tV}, \textbf{-bV}, and \textbf{-gV}. There is also a substantial number of non-gradating forms where the activation for 484 is $> 0.5$. Most of these fall into the \textbf{linnoitus} `fortress' / \textbf{linnoituksen} `fortress sg. gen.' patterns where a {\bf k} is inserted in the penultimate syllable. This alternation bears great resemblance to gradation as mentioned in Section \ref{sec:cg}. There are also a few examples of the type \textbf{tase} `balance sheet' / \textbf{taseen} `balance sheet sg. gen.' where the stem-final vowel is doubled displaying large activation for 484. This is perhaps somewhat harder to explain. However, note that this vowel doubling frequently co-occurs with gradation as in \textbf{tarvi\underline{k}e} `accessory', \textbf{tarvi\underline{kk}een} `accessory sg. gen.'.

In our experiments we found that the system would sometimes output a gradated form even when the exact type of gradation was not present in the training data, for example {\bf ba\underline{mb}u} $\sim$ {\bf ba\underline{mm}un} (`bamboo' sg. nom. $\sim$ sg. gen.).  Since Finnish natively lacks {\bf b} and {\bf g}, examples of gradation with these consonants are rare. However, it is indeed the case that loanwords that include such voiced stops do undergo gradation, e.g. {\bf du\underline{b}ata} $\sim$ {\bf du\underline{bb}aan} (`to dub' inf. $\sim$ 1p sg. pres. sg.) \cite{voutilainen:08}. Since native Finnish speakers seem to extend gradation from voiceless stops to their voiced counterparts in loanwords, the question whether neural models can exhibit such generalizing behavior as well is an interesting one. Our initial investigations into whether the similarity of the learned embeddings for {\bf p} and {\bf b} could trigger such generalizations across similar sounds failed to identify a clear reason for the behavior, and we leave a detailed study of this to future work.

We have presented an investigation of encoder representations of phonological alternations, specifically consonant gradation in Finnish. We found evidence of a generalized representation of gradation covering all stops which undergo gradation and different types of gradation. We also found that scaling hidden states can ``switch off" gradation, prompting the model to generate alternate forms which do not display gradation. Moreover, the direction of gradation can be encoded as positive vs. negative hidden dimension activation.

\bibliography{eacl2021}

\begin{thebibliography}{26}
\expandafter\ifx\csname natexlab\endcsname\relax\def\natexlab#1{#1}\fi

\bibitem[{Abondolo(1998)}]{abondolo1998}
Daniel Abondolo. 1998.
\newblock Finnish.
\newblock In Daniel Abondolo, editor, \emph{The Uralic Languages}. Routledge.

\bibitem[{Bahdanau et~al.(2014)Bahdanau, Cho, and Bengio}]{bahdanau2014neural}
Dzmitry Bahdanau, Kyunghyun Cho, and Yoshua Bengio. 2014.
\newblock Neural machine translation by jointly learning to align and
  translate.
\newblock \emph{arXiv preprint arXiv:1409.0473}.

\bibitem[{Begu{\v{s}}(2020{\natexlab{a}})}]{beguvs2020identity}
Ga{\v{s}}per Begu{\v{s}}. 2020{\natexlab{a}}.
\newblock Identity-based patterns in deep convolutional networks: Generative
  adversarial phonology and reduplication.
\newblock \emph{arXiv preprint arXiv:2009.06110}.

\bibitem[{Begu{\v{s}}(2020{\natexlab{b}})}]{beguvs2020modeling}
Ga{\v{s}}per Begu{\v{s}}. 2020{\natexlab{b}}.
\newblock Modeling unsupervised phonetic and phonological learning in
  generative adversarial phonology.
\newblock \emph{Proceedings of the Society for Computation in Linguistics},
  3(1):138--148.

\bibitem[{Clark et~al.(2019)Clark, Khandelwal, Levy, and
  Manning}]{clark2019does}
Kevin Clark, Urvashi Khandelwal, Omer Levy, and Christopher~D. Manning. 2019.
\newblock \href {https://doi.org/10.18653/v1/W19-4828} {What does {BERT} look
  at? {A}n analysis of {BERT}{'}s attention}.
\newblock In \emph{Proceedings of the 2019 ACL Workshop BlackboxNLP: Analyzing
  and Interpreting Neural Networks for NLP}, pages 276--286, Florence, Italy.
  Association for Computational Linguistics.

\bibitem[{Conneau et~al.(2018{\natexlab{a}})Conneau, Kruszewski, Lample,
  Barrault, and Baroni}]{conneau2018you}
Alexis Conneau, German Kruszewski, Guillaume Lample, Lo{\"\i}c Barrault, and
  Marco Baroni. 2018{\natexlab{a}}.
\newblock \href {https://doi.org/10.18653/v1/P18-1198} {What you can cram into
  a single {\$}{\&}!{\#}* vector: Probing sentence embeddings for linguistic
  properties}.
\newblock In \emph{Proceedings of the 56th Annual Meeting of the Association
  for Computational Linguistics (Volume 1: Long Papers)}, pages 2126--2136,
  Melbourne, Australia. Association for Computational Linguistics.

\bibitem[{Conneau et~al.(2018{\natexlab{b}})Conneau, Rinott, Lample, Williams,
  Bowman, Schwenk, and Stoyanov}]{conneau2018xnli}
Alexis Conneau, Ruty Rinott, Guillaume Lample, Adina Williams, Samuel Bowman,
  Holger Schwenk, and Veselin Stoyanov. 2018{\natexlab{b}}.
\newblock \href {https://doi.org/10.18653/v1/D18-1269} {{XNLI}: Evaluating
  cross-lingual sentence representations}.
\newblock In \emph{Proceedings of the 2018 Conference on Empirical Methods in
  Natural Language Processing}, pages 2475--2485, Brussels, Belgium.
  Association for Computational Linguistics.

\bibitem[{Cotterell et~al.(2017)Cotterell, Kirov, Sylak-Glassman, Walther,
  Vylomova, Xia, Faruqui, K{\"u}bler, Yarowsky, Eisner, and
  Hulden}]{cotterell2017conll}
Ryan Cotterell, Christo Kirov, John Sylak-Glassman, G{\'e}raldine Walther,
  Ekaterina Vylomova, Patrick Xia, Manaal Faruqui, Sandra K{\"u}bler, David
  Yarowsky, Jason Eisner, and Mans Hulden. 2017.
\newblock \href {https://doi.org/10.18653/v1/K17-2001} {{C}o{NLL}-{SIGMORPHON}
  2017 shared task: Universal morphological reinflection in 52 languages}.
\newblock In \emph{Proceedings of the {C}o{NLL} {SIGMORPHON} 2017 Shared Task:
  Universal Morphological Reinflection}, pages 1--30, Vancouver. Association
  for Computational Linguistics.

\bibitem[{Cotterell et~al.(2016)Cotterell, Kirov, Sylak-Glassman, Yarowsky,
  Eisner, and Hulden}]{cotterell2016sigmorphon}
Ryan Cotterell, Christo Kirov, John Sylak-Glassman, David Yarowsky, Jason
  Eisner, and Mans Hulden. 2016.
\newblock The {SIGMORPHON} 2016 shared task—morphological reinflection.
\newblock In \emph{Proceedings of the 14th SIGMORPHON Workshop on Computational
  Research in Phonetics, Phonology, and Morphology}, pages 10--22.

\bibitem[{Dalvi et~al.(2019)Dalvi, Durrani, Sajjad, Belinkov, Bau, and
  Glass}]{dalvi2019one}
Fahim Dalvi, Nadir Durrani, Hassan Sajjad, Yonatan Belinkov, Anthony Bau, and
  James Glass. 2019.
\newblock What is one grain of sand in the desert? {A}nalyzing individual
  neurons in deep {NLP} models.
\newblock In \emph{Proceedings of the AAAI Conference on Artificial
  Intelligence}, volume~33, pages 6309--6317.

\bibitem[{Doshi-Velez and Kim(2017)}]{doshi2017towards}
Finale Doshi-Velez and Been Kim. 2017.
\newblock Towards a rigorous science of interpretable machine learning.
\newblock \emph{arXiv preprint arXiv:1702.08608}.

\bibitem[{Elman(1990)}]{elman1990}
Jeffrey~L. Elman. 1990.
\newblock Finding structure in time.
\newblock \emph{Cognitive Science}, 14(2):179--211.

\bibitem[{Goodfellow et~al.(2014)Goodfellow, Pouget-Abadie, Mirza, Xu,
  Warde-Farley, Ozair, Courville, and Bengio}]{goodfellow2014generative}
Ian~J. Goodfellow, Jean Pouget-Abadie, Mehdi Mirza, Bing Xu, David
  Warde-Farley, Sherjil Ozair, Aaron Courville, and Yoshua Bengio. 2014.
\newblock \href {http://arxiv.org/abs/1406.2661} {Generative adversarial
  networks}.

\bibitem[{Hakulinen et~al.(2004)Hakulinen, Vilkuna, Korhonen, Koivisto,
  Heinonen, and Alho}]{visk2004iso}
Auli Hakulinen, Maria Vilkuna, Riitta Korhonen, Vesa Koivisto, Tarja-Riitta
  Heinonen, and Irja Alho. 2004.
\newblock \emph{Iso Suomen Kielioppi [{C}omprehensive grammar of {F}innish]}.
\newblock Suomalaisen Kirjallisuuden Seura [Finnish Literature Society].

\bibitem[{Haverinen et~al.(2014)Haverinen, Nyblom, Viljanen, Laippala, Kohonen,
  Missil{\"a}, Ojala, Salakoski, and Ginter}]{haverinen:13}
Katri Haverinen, Jenna Nyblom, Timo Viljanen, Veronika Laippala, Samuel
  Kohonen, Anna Missil{\"a}, Stina Ojala, Tapio Salakoski, and Filip Ginter.
  2014.
\newblock Building the essential resources for {F}innish: the {T}urku
  {D}ependency {T}reebank.
\newblock \emph{Language Resources and Evaluation}, 48(3):493--531.

\bibitem[{Kann and Sch{\"u}tze(2016)}]{kann2016med}
Katharina Kann and Hinrich Sch{\"u}tze. 2016.
\newblock {MED}: The {LMU} system for the {SIGMORPHON} 2016 shared task on
  morphological reinflection.
\newblock In \emph{Proceedings of the 14th SIGMORPHON Workshop on Computational
  Research in Phonetics, Phonology, and Morphology}, pages 62--70.

\bibitem[{Karpathy et~al.(2015)Karpathy, Johnson, and
  Fei-Fei}]{karpathy2015visualizing}
Andrej Karpathy, Justin Johnson, and Li~Fei-Fei. 2015.
\newblock Visualizing and understanding recurrent networks.
\newblock \emph{arXiv preprint arXiv:1506.02078}.

\bibitem[{Klein et~al.(2018)Klein, Kim, Deng, Nguyen, Senellart, and
  Rush}]{klein-etal-2018-opennmt}
Guillaume Klein, Yoon Kim, Yuntian Deng, Vincent Nguyen, Jean Senellart, and
  Alexander Rush. 2018.
\newblock \href {https://www.aclweb.org/anthology/W18-1817} {{O}pen{NMT}:
  Neural machine translation toolkit}.
\newblock In \emph{Proceedings of the 13th Conference of the Association for
  Machine Translation in the {A}mericas (Volume 1: Research Track)}, pages
  177--184, Boston, MA. Association for Machine Translation in the Americas.

\bibitem[{Kolachina and Magyar(2019)}]{kolachina2019phone}
Sudheer Kolachina and Lilla Magyar. 2019.
\newblock What do phone embeddings learn about phonology?
\newblock In \emph{Proceedings of the 16th Workshop on Computational Research
  in Phonetics, Phonology, and Morphology}, pages 160--169.

\bibitem[{Meyes et~al.(2019)Meyes, Lu, de~Puiseau, and
  Meisen}]{meyes2019ablation}
R.~Meyes, M.~Lu, C.~Waubert de~Puiseau, and T.~Meisen. 2019.
\newblock Ablation studies to uncover structure of learned representations in
  artificial neural networks.
\newblock In \emph{Proceedings on the International Conference on Artificial
  Intelligence (ICAI)}, pages 185--191. The Steering Committee of The World
  Congress in Computer Science.

\bibitem[{Mikolov et~al.(2013)Mikolov, Sutskever, Chen, Corrado, and
  Dean}]{mikolov2013distributed}
Tomas Mikolov, Ilya Sutskever, Kai Chen, Greg~S. Corrado, and Jeff Dean. 2013.
\newblock Distributed representations of words and phrases and their
  compositionality.
\newblock In \emph{Advances in neural information processing systems}, pages
  3111--3119.

\bibitem[{Pirinen(2015)}]{pirinen2015development}
Tommi~A. Pirinen. 2015.
\newblock Development and use of computational morphology of {F}innish in the
  open source and open science era: Notes on experiences with {O}morfi
  development.
\newblock \emph{SKY Journal of Linguistics}, 28:381--393.

\bibitem[{Rodd(1997)}]{rodd1997recurrent}
Jennifer Rodd. 1997.
\newblock Recurrent neural-network learning of phonological regularities in
  {T}urkish.
\newblock In \emph{CoNLL97: Computational Natural Language Learning}.

\bibitem[{Silfverberg et~al.(2018)Silfverberg, Mao, and
  Hulden}]{silfverberg-etal-2018-sound}
Miikka~P. Silfverberg, Lingshuang Mao, and Mans Hulden. 2018.
\newblock \href {https://doi.org/10.7275/R5NZ85VD} {Sound analogies with
  phoneme embeddings}.
\newblock In \emph{Proceedings of the Society for Computation in Linguistics
  ({SC}i{L}) 2018}, pages 136--144.

\bibitem[{Torroba~Hennigen et~al.(2020)Torroba~Hennigen, Williams, and
  Cotterell}]{hennigen2020intrinsic}
Lucas Torroba~Hennigen, Adina Williams, and Ryan Cotterell. 2020.
\newblock \href {https://doi.org/10.18653/v1/2020.emnlp-main.15} {Intrinsic
  probing through dimension selection}.
\newblock In \emph{Proceedings of the 2020 Conference on Empirical Methods in
  Natural Language Processing (EMNLP)}, pages 197--216, Online. Association for
  Computational Linguistics.

\bibitem[{Voutilainen(2008)}]{voutilainen:08}
Eero Voutilainen. 2008.
\newblock Bloggaaminen laajentaa astevaihtelua [{B}logging extends consonant
  gradation].
\newblock
  \url{https://www.kotus.fi/nyt/kolumnit_artikkelit_ja_esitelmat/kieli-ikkuna_(1996_2010)/bloggaaminen_laajentaa_astevaihtelua}.

\end{thebibliography}
\bibliographystyle{acl_natbib}
\newpage
$ $
\newpage
\appendix

\section{Appendix}
\label{sec:appendix}

\begin{figure*}[!htb]
\includegraphics[width=0.45\textwidth]{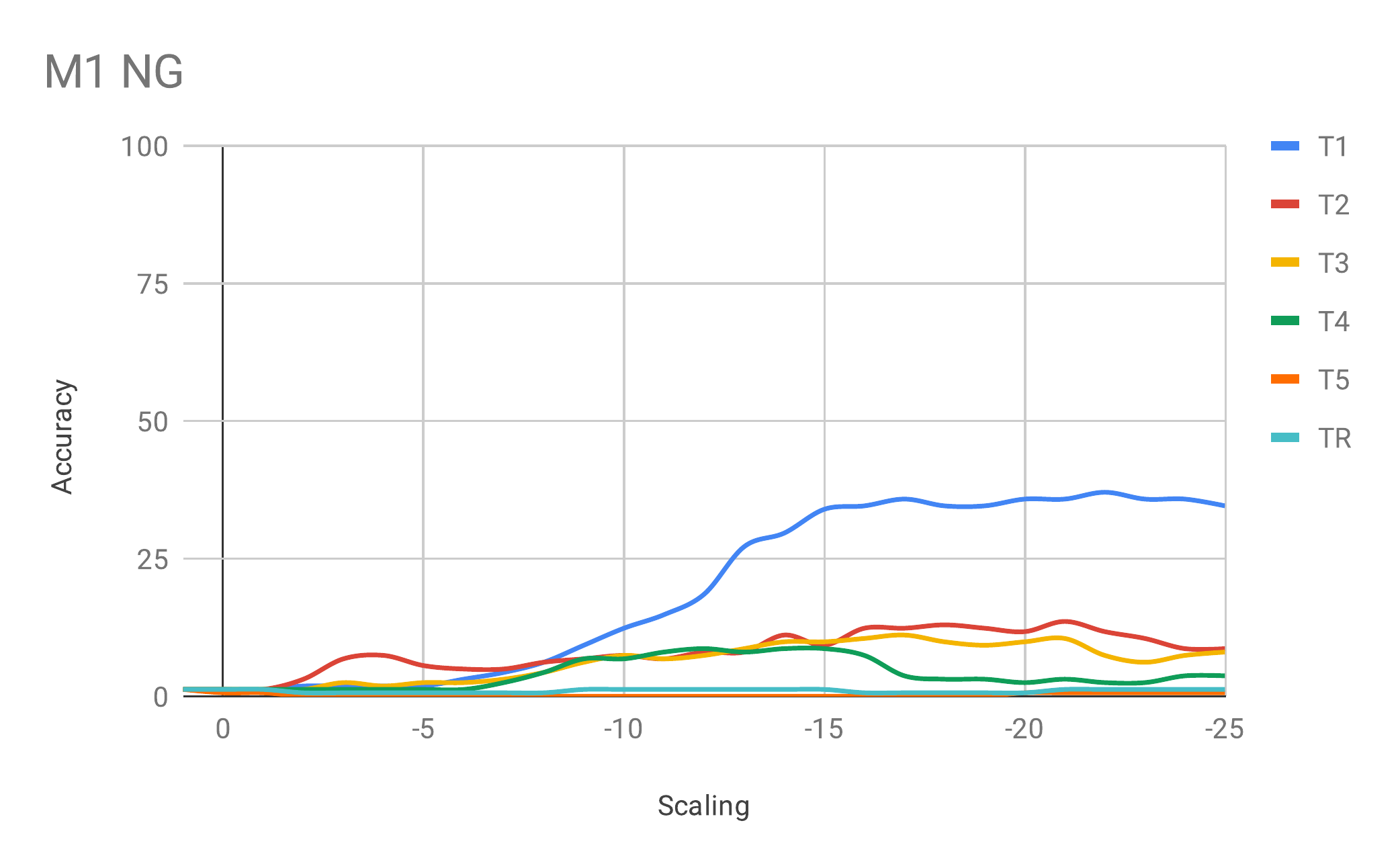}~~~\includegraphics[width=0.45\textwidth]{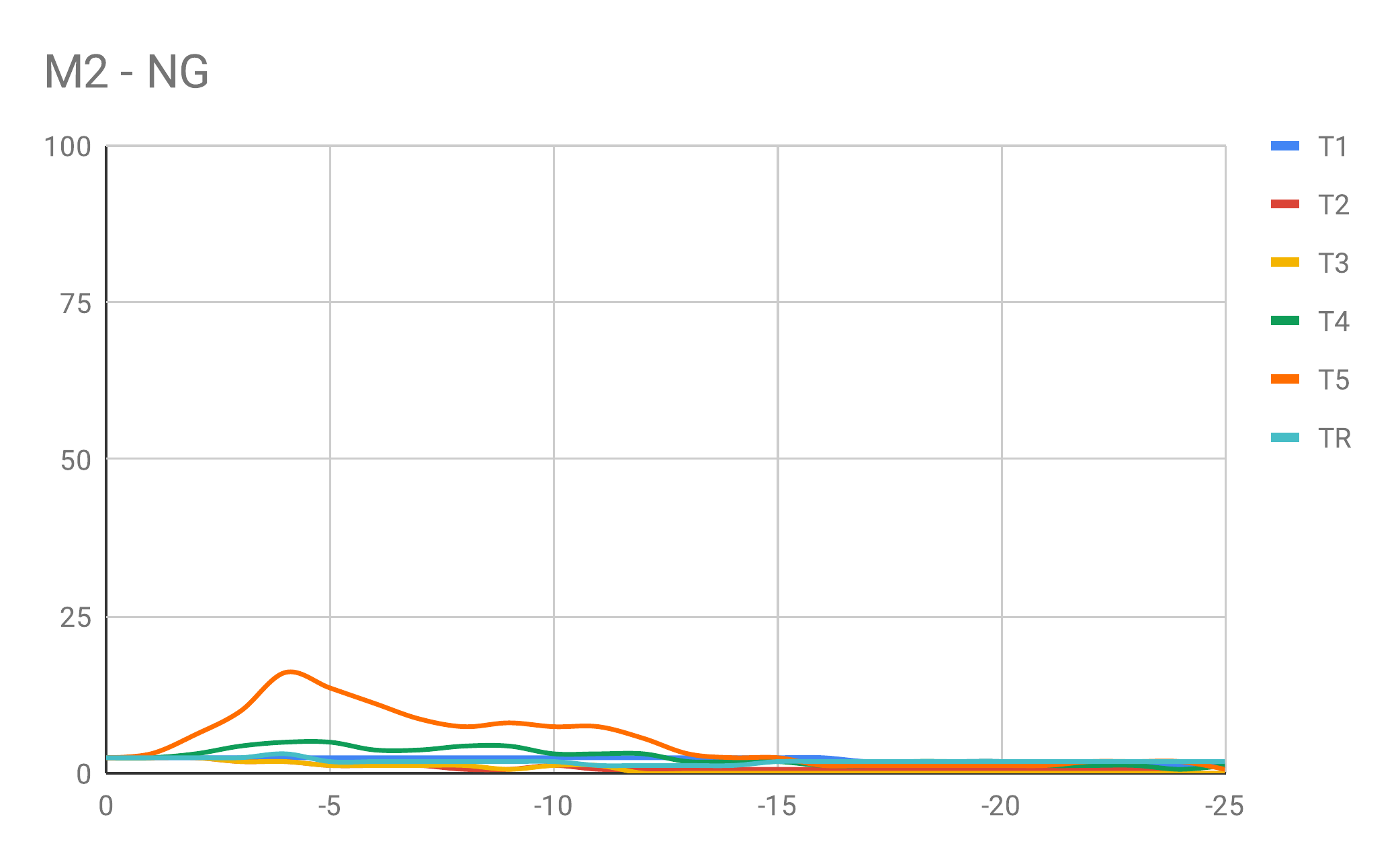}\\
\includegraphics[width=0.45\textwidth]{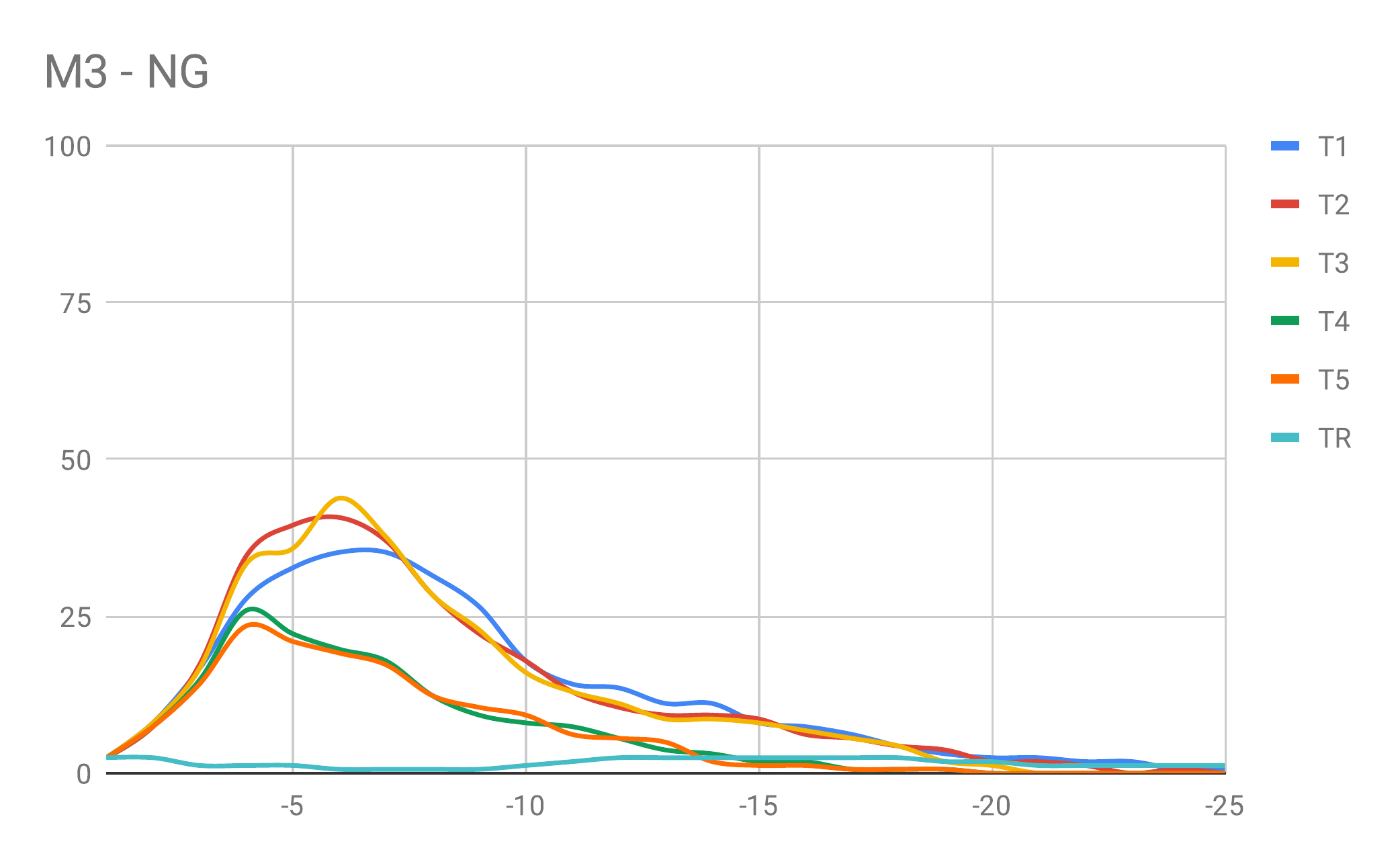}~~~\includegraphics[width=0.45\textwidth]{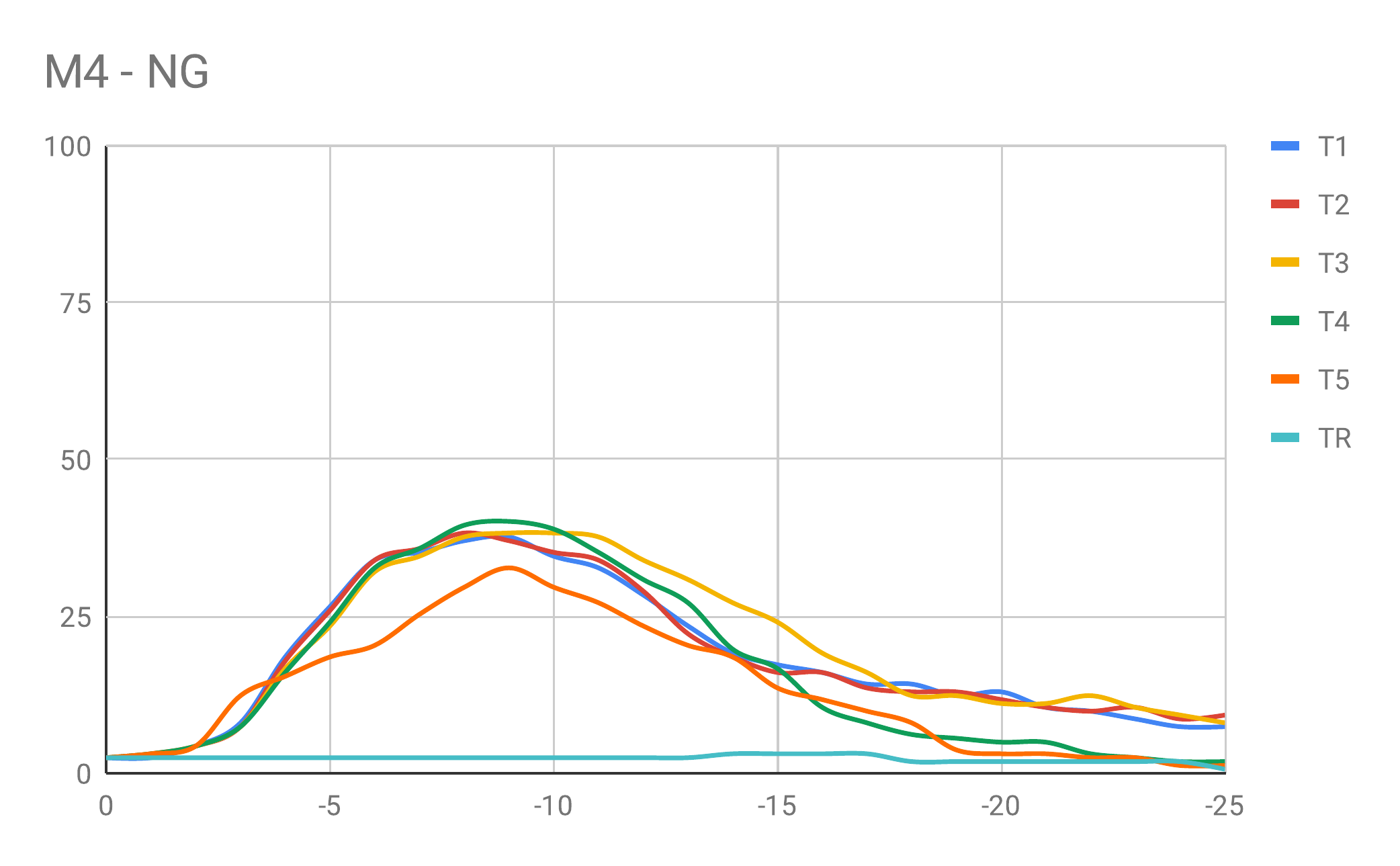}\\
\includegraphics[width=0.45\textwidth]{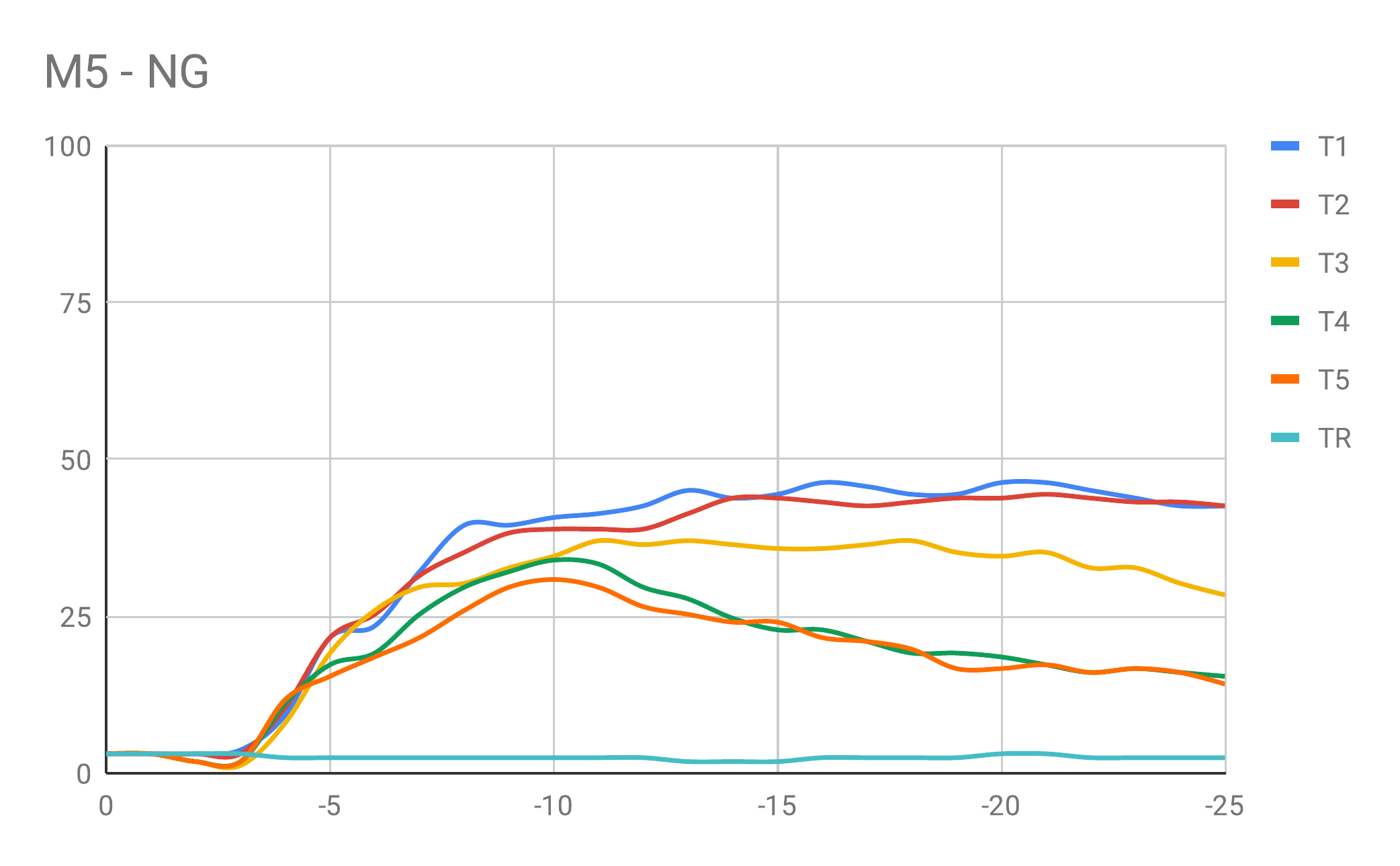}~~~\includegraphics[width=0.45\textwidth]{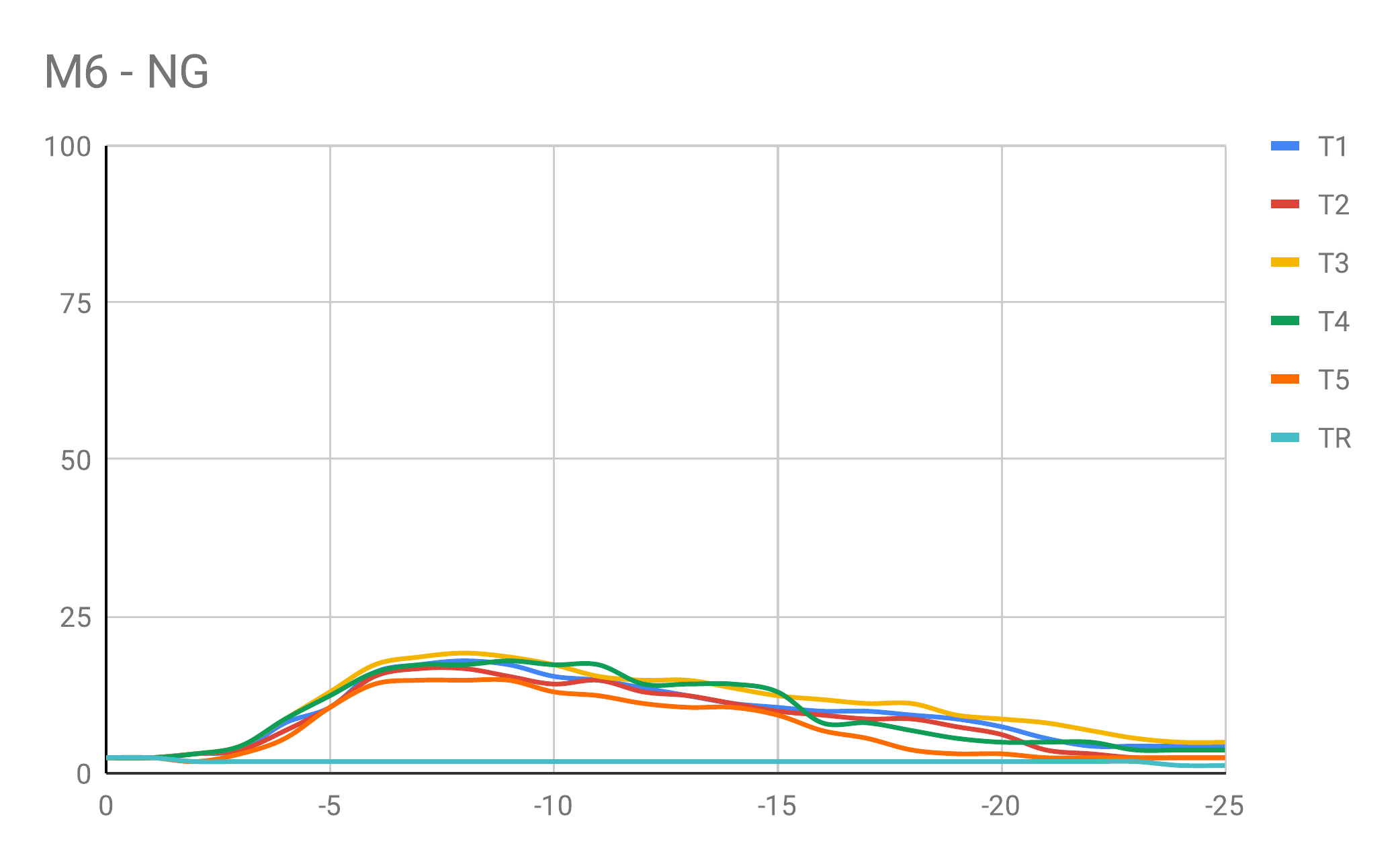}\\
\includegraphics[width=0.45\textwidth]{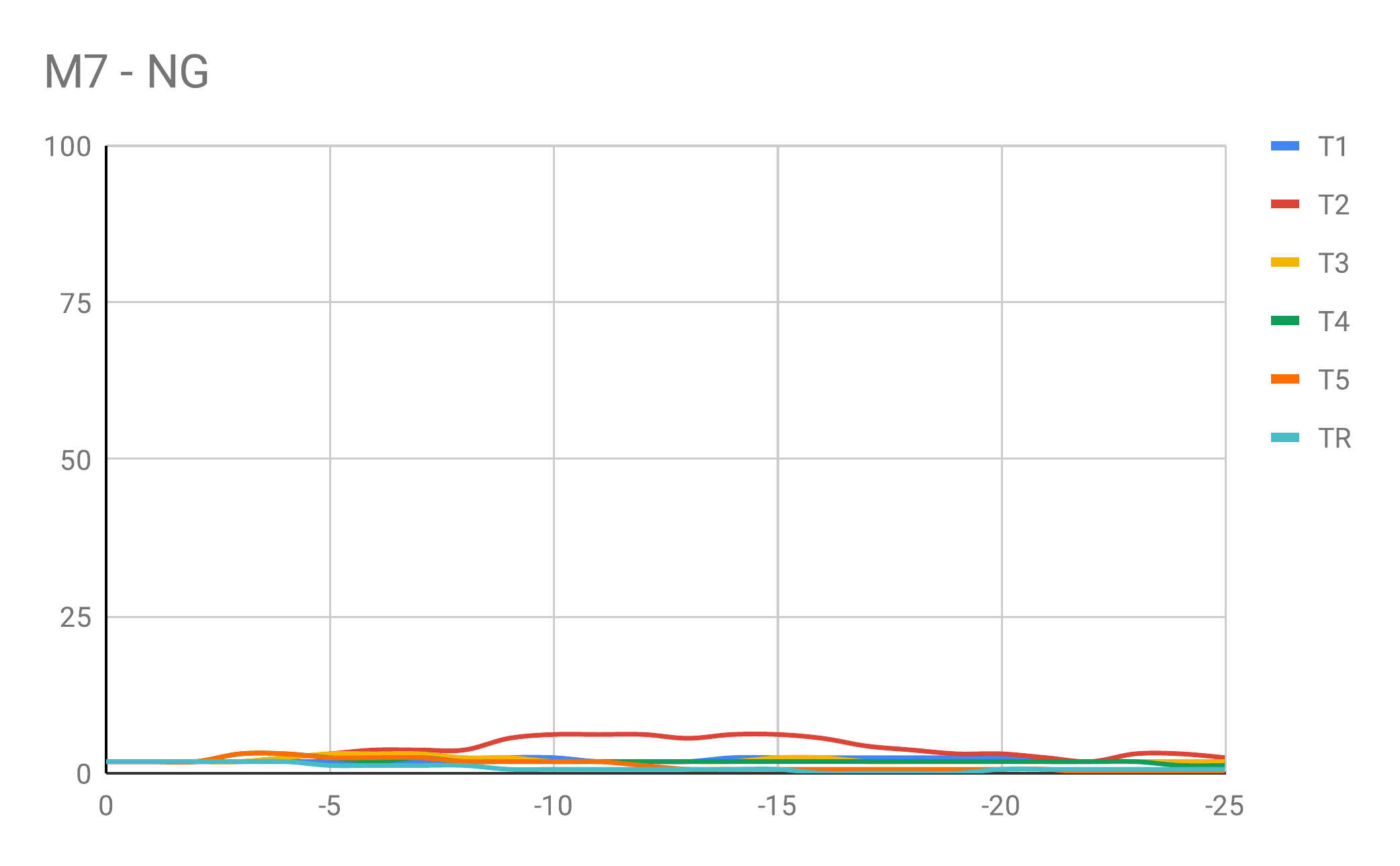}~~~\includegraphics[width=0.45\textwidth]{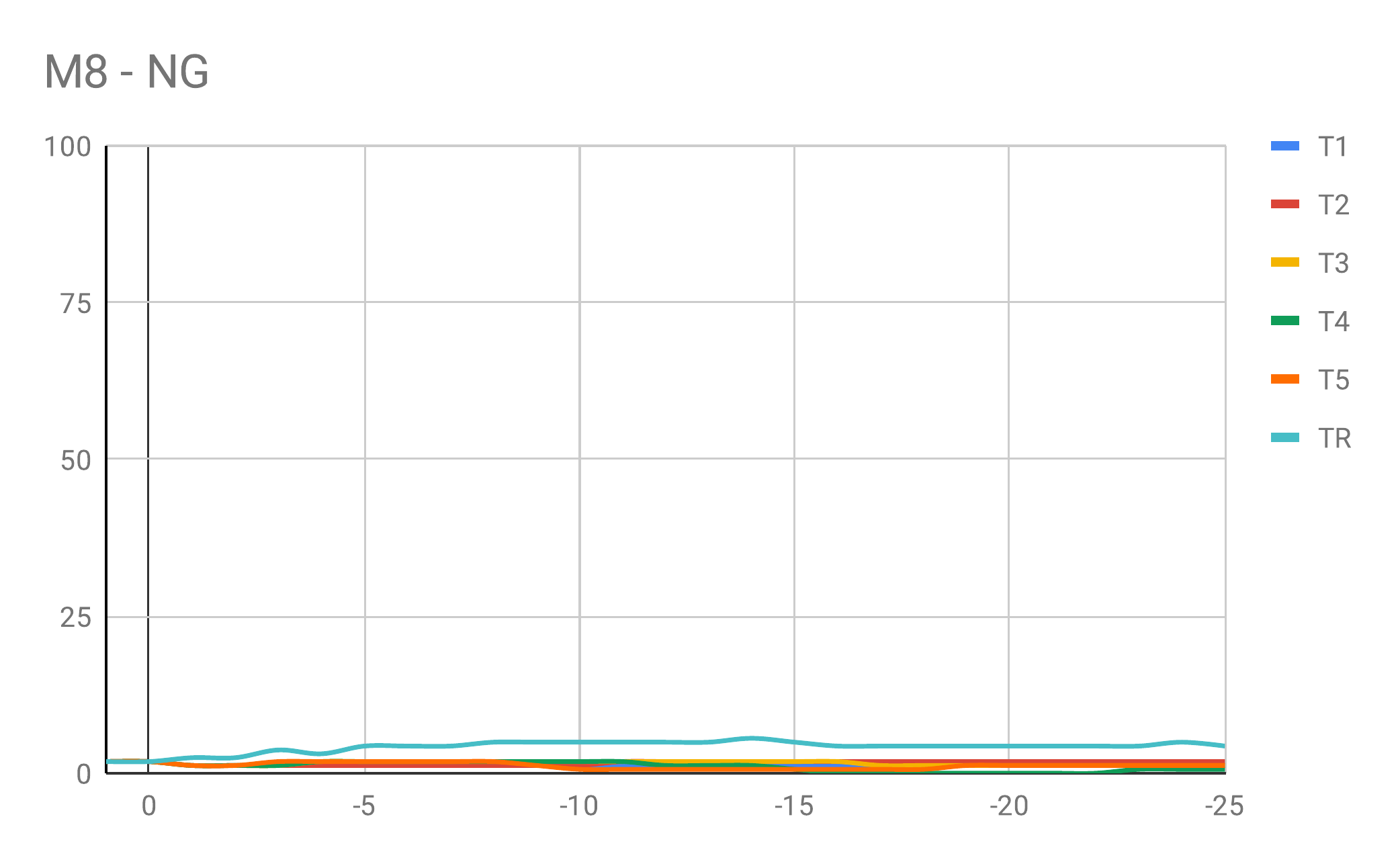}\\
\includegraphics[width=0.45\textwidth]{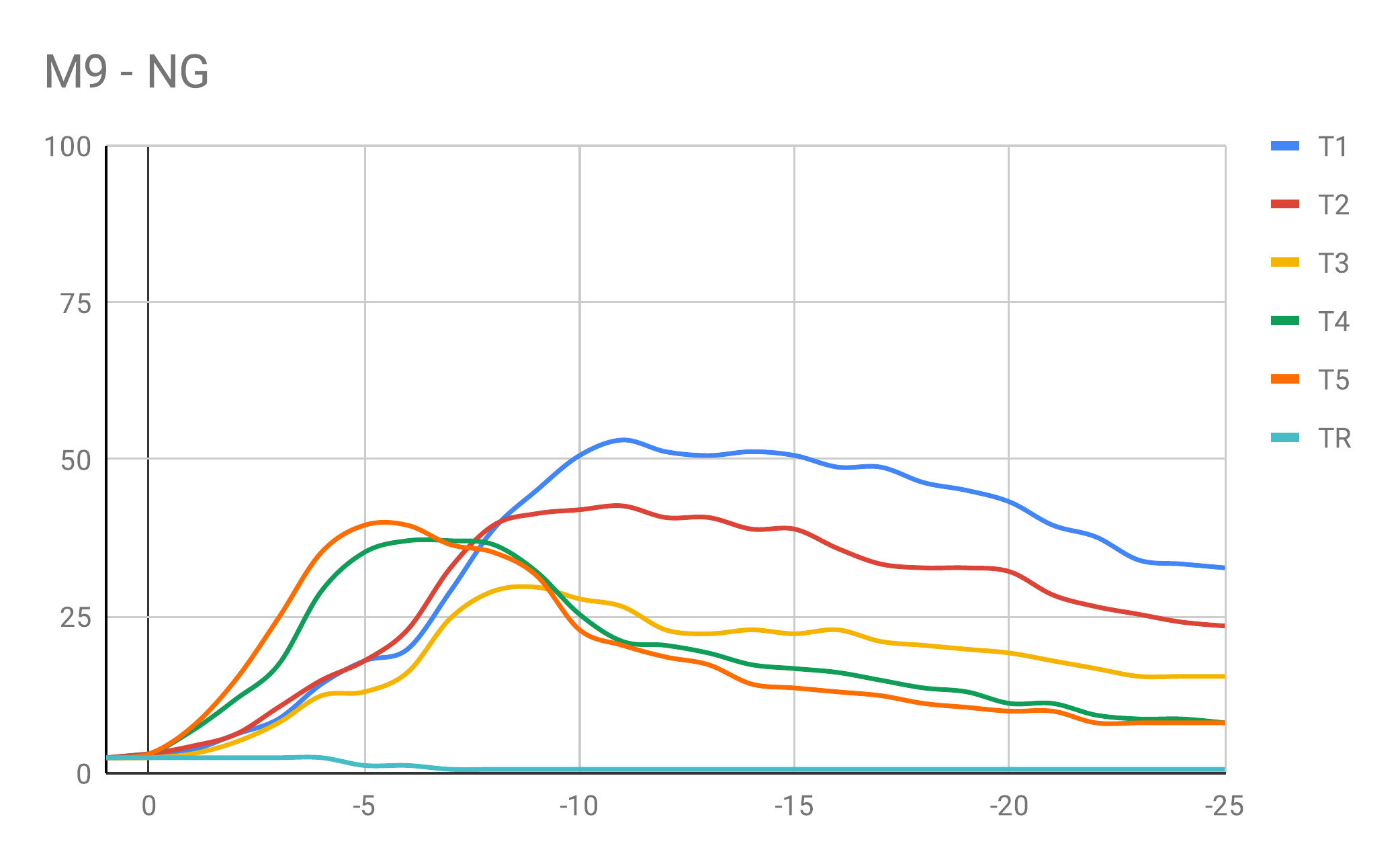}~~~\includegraphics[width=0.45\textwidth]{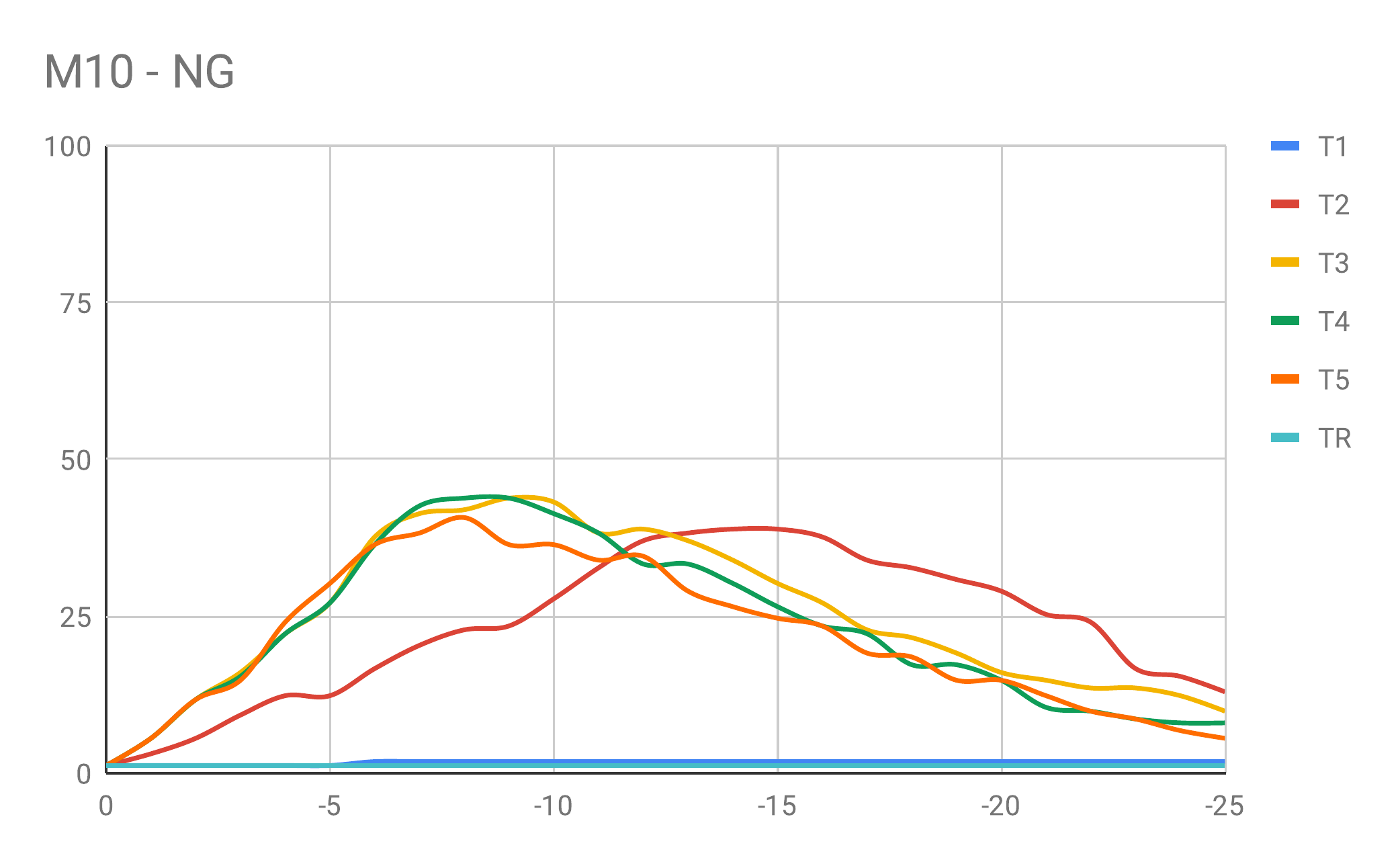}\\
\caption{Results when scaling the activations for the top 1-5 (T1 - T5) encoder dimensions associated with gradation for each of out ten models M1 - M10. These graphs show the amount of outputs which do not undergo gradation which are produced when encoder dimensions are scaled. As comparison, the green TR graph shows the effect of scaling 5 randomly selected encoder dimensions.}
\label{fig:all-ng-scaling}
\end{figure*}

\begin{figure*}[!htb]
\includegraphics[width=0.45\textwidth]{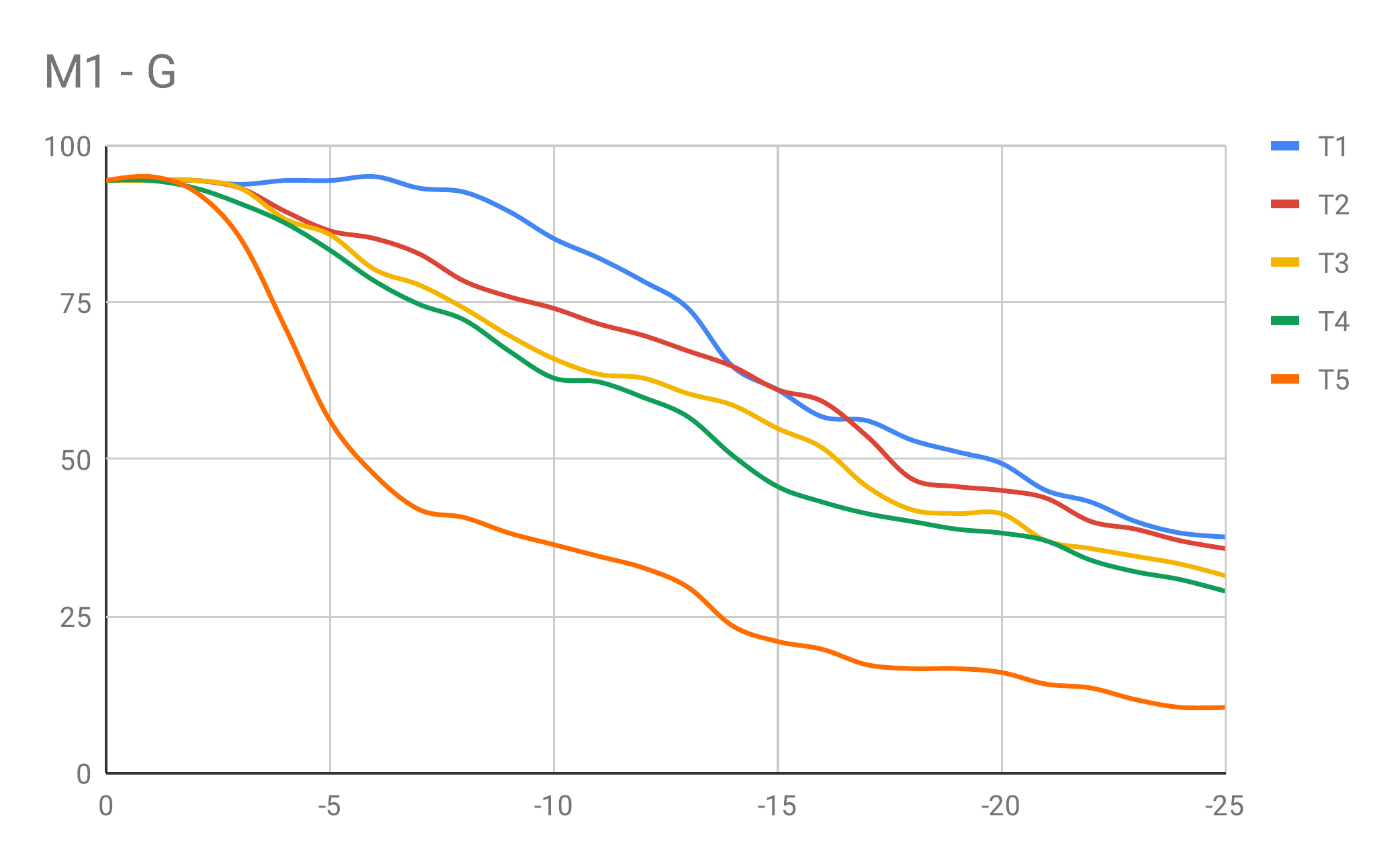}~~~\includegraphics[width=0.45\textwidth]{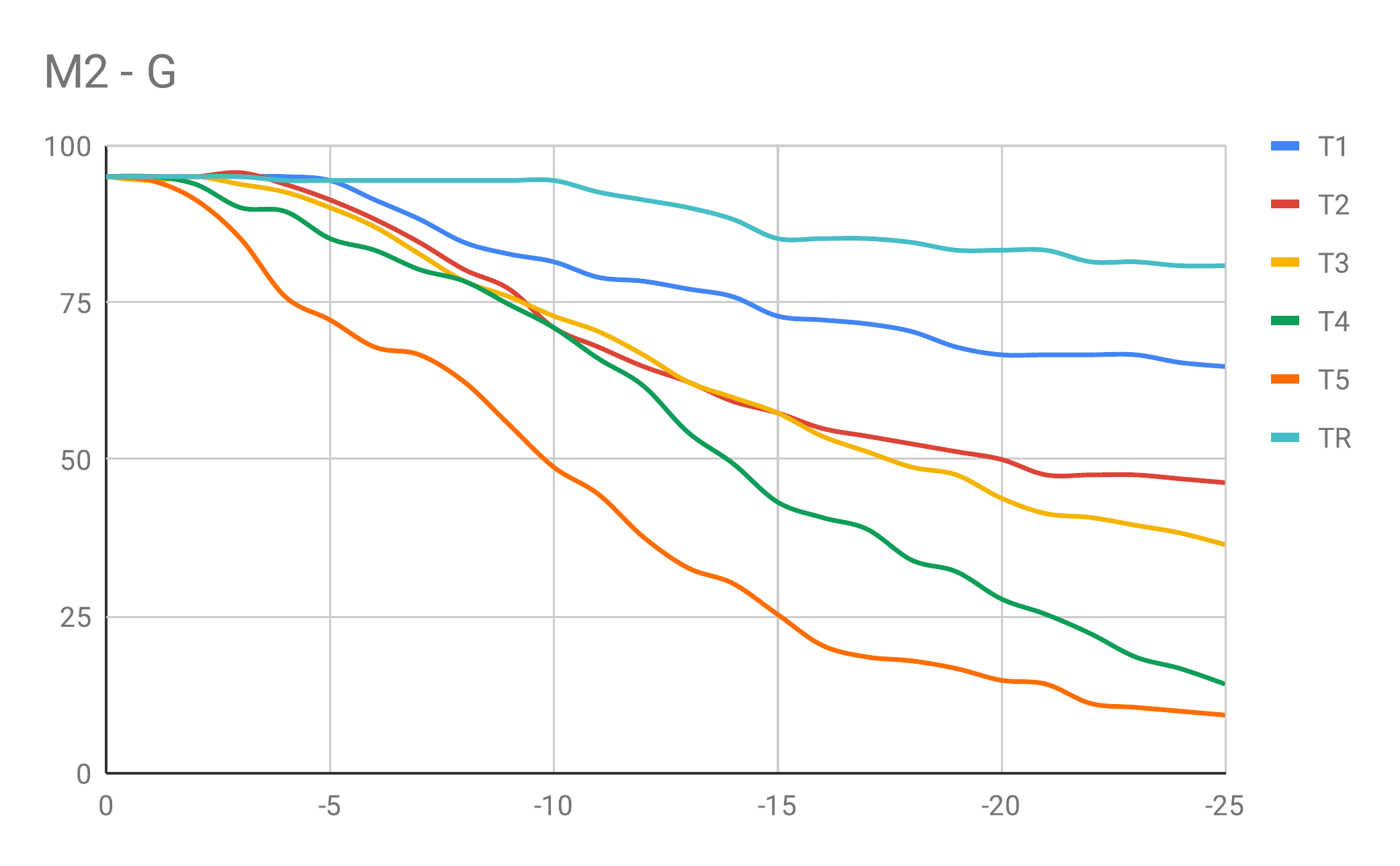}\\
\includegraphics[width=0.45\textwidth]{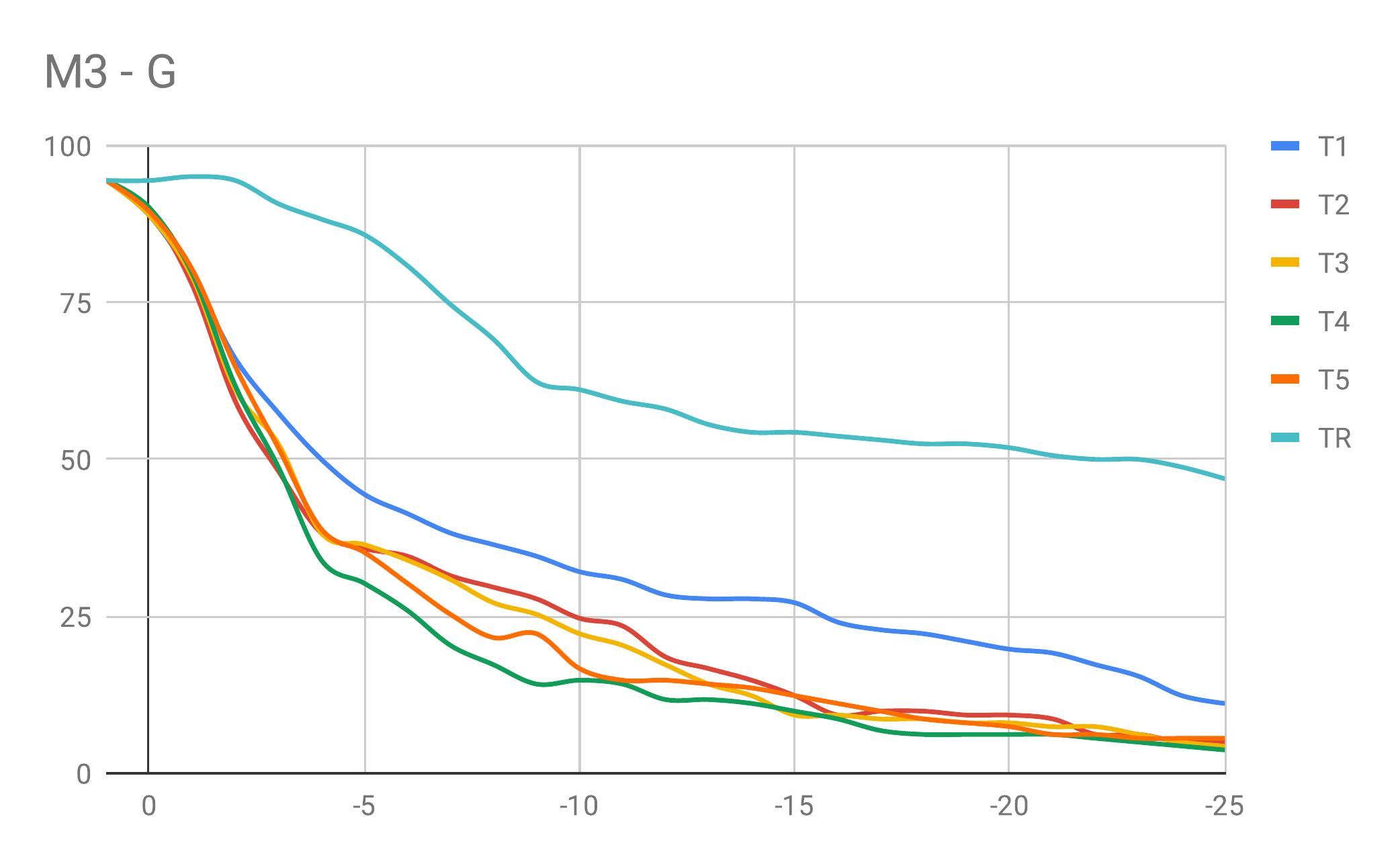}~~~\includegraphics[width=0.45\textwidth]{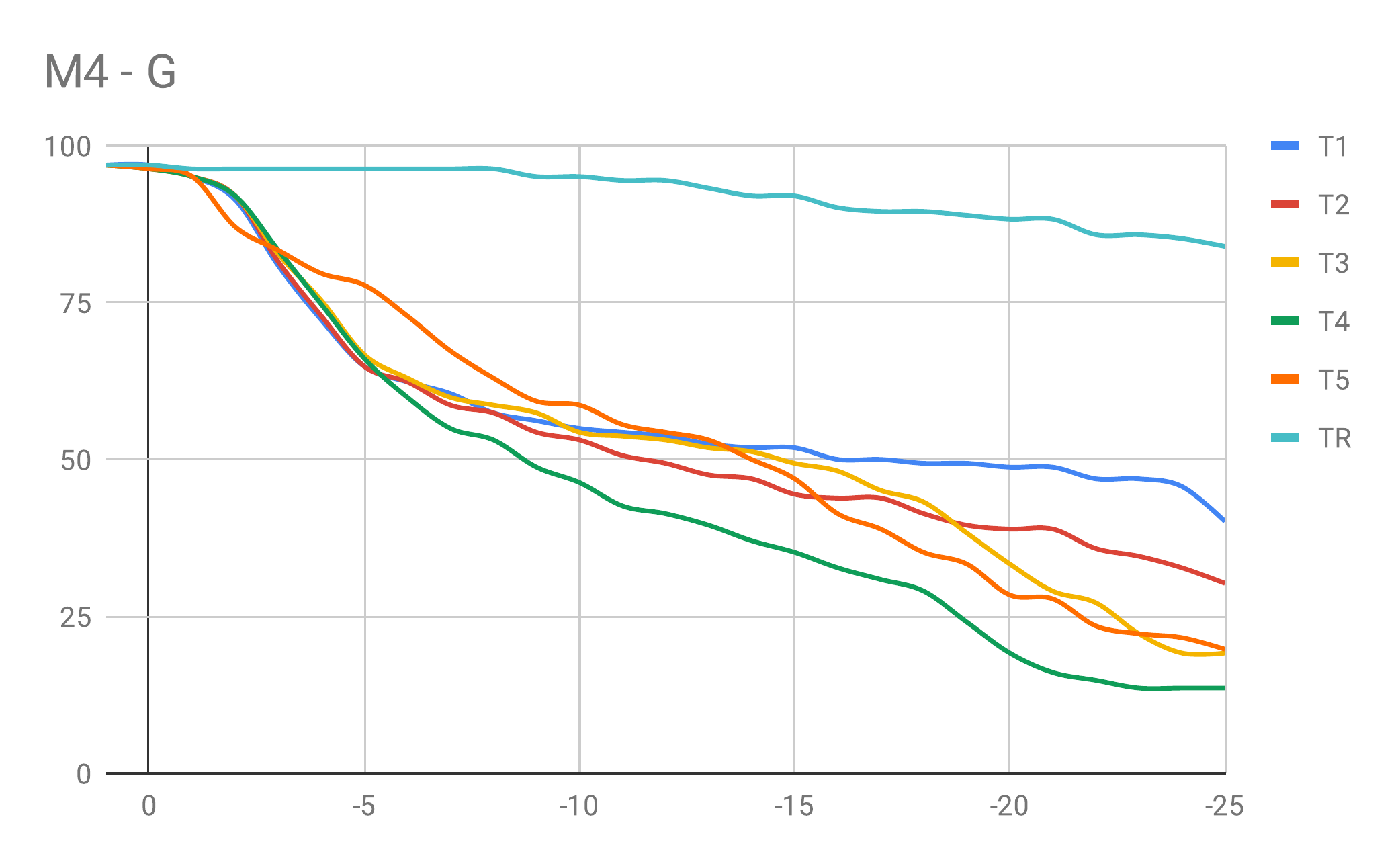}\\
\includegraphics[width=0.45\textwidth]{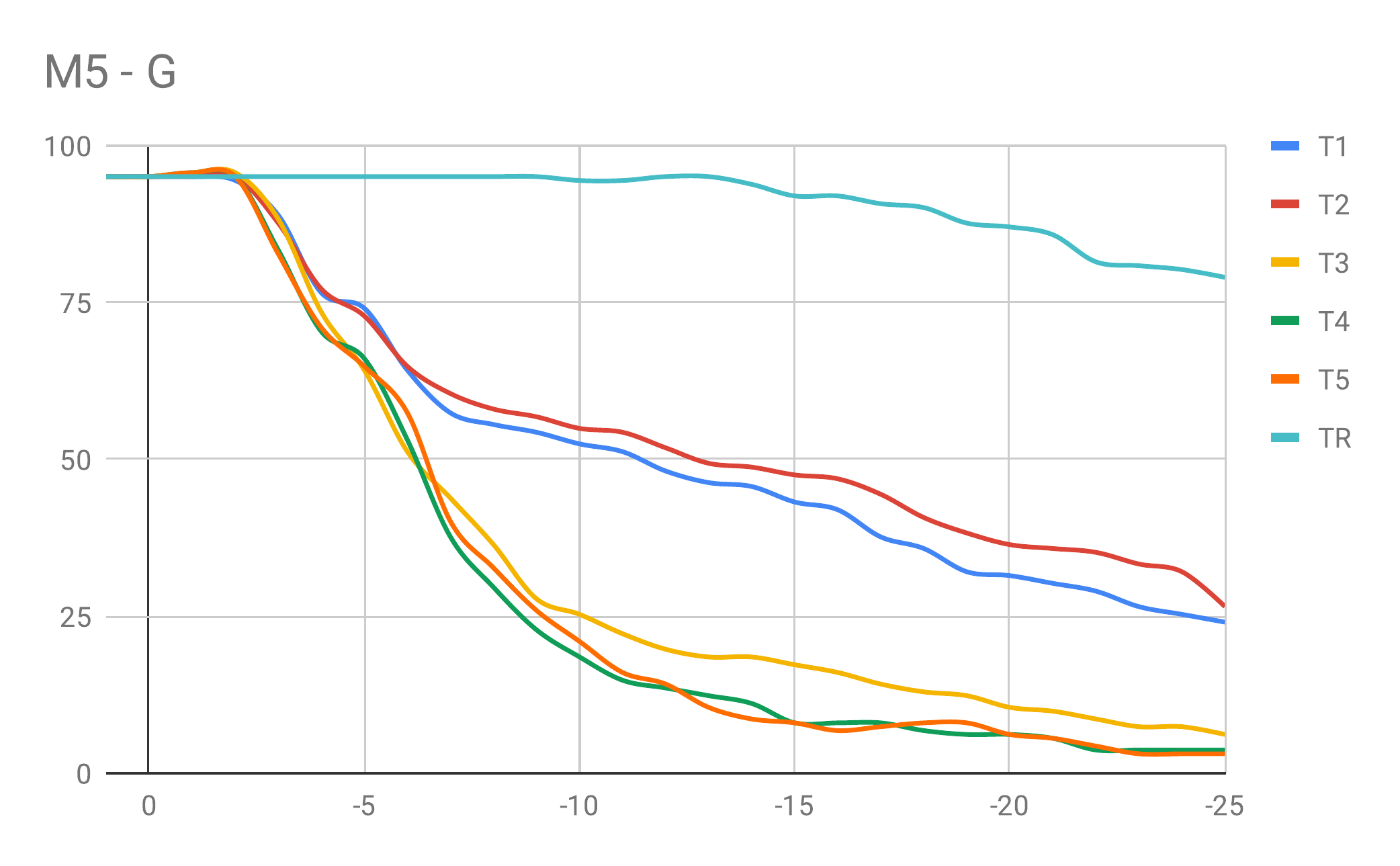}~~~\includegraphics[width=0.45\textwidth]{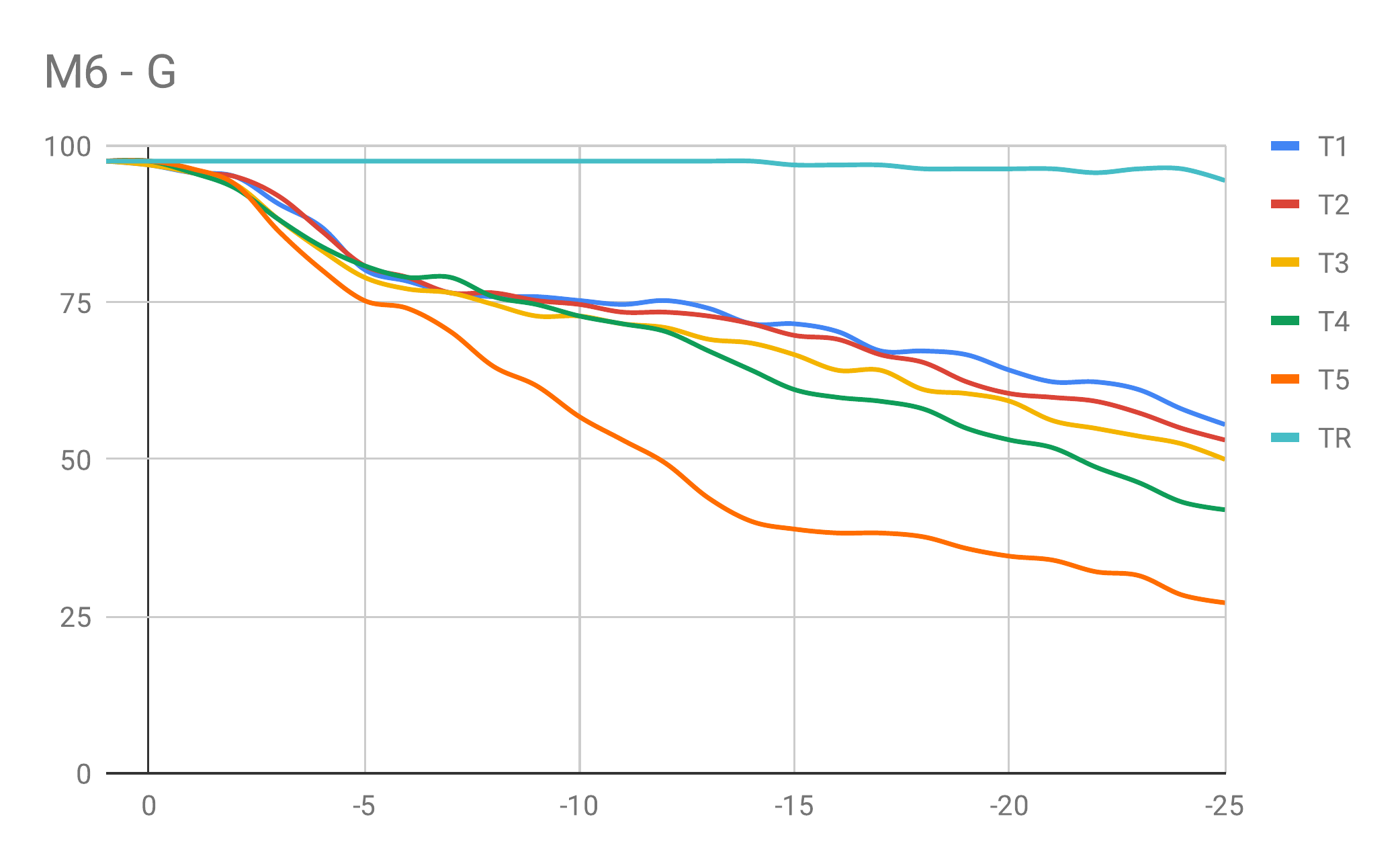}\\
\includegraphics[width=0.45\textwidth]{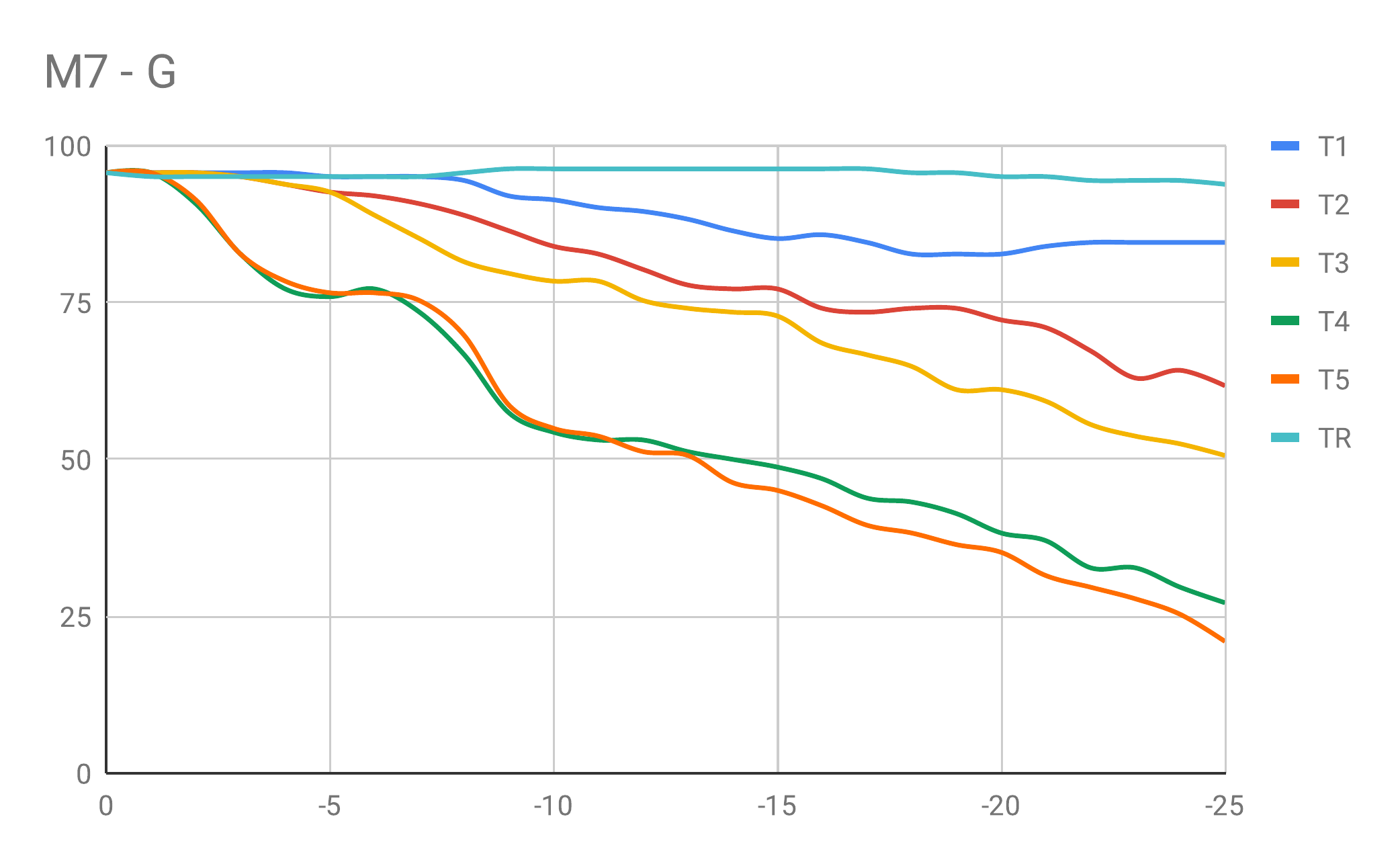}~~~\includegraphics[width=0.45\textwidth]{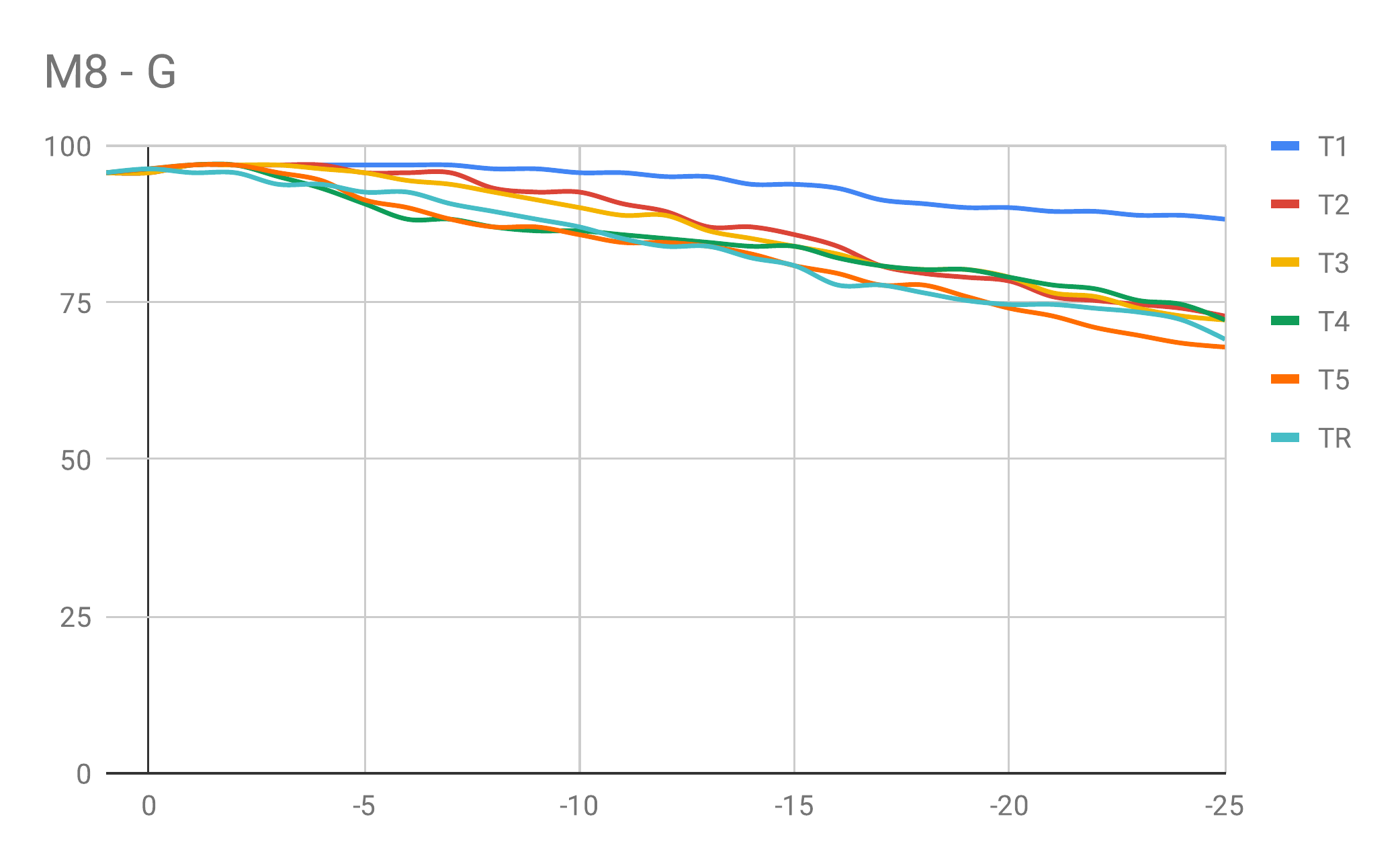}\\
\includegraphics[width=0.45\textwidth]{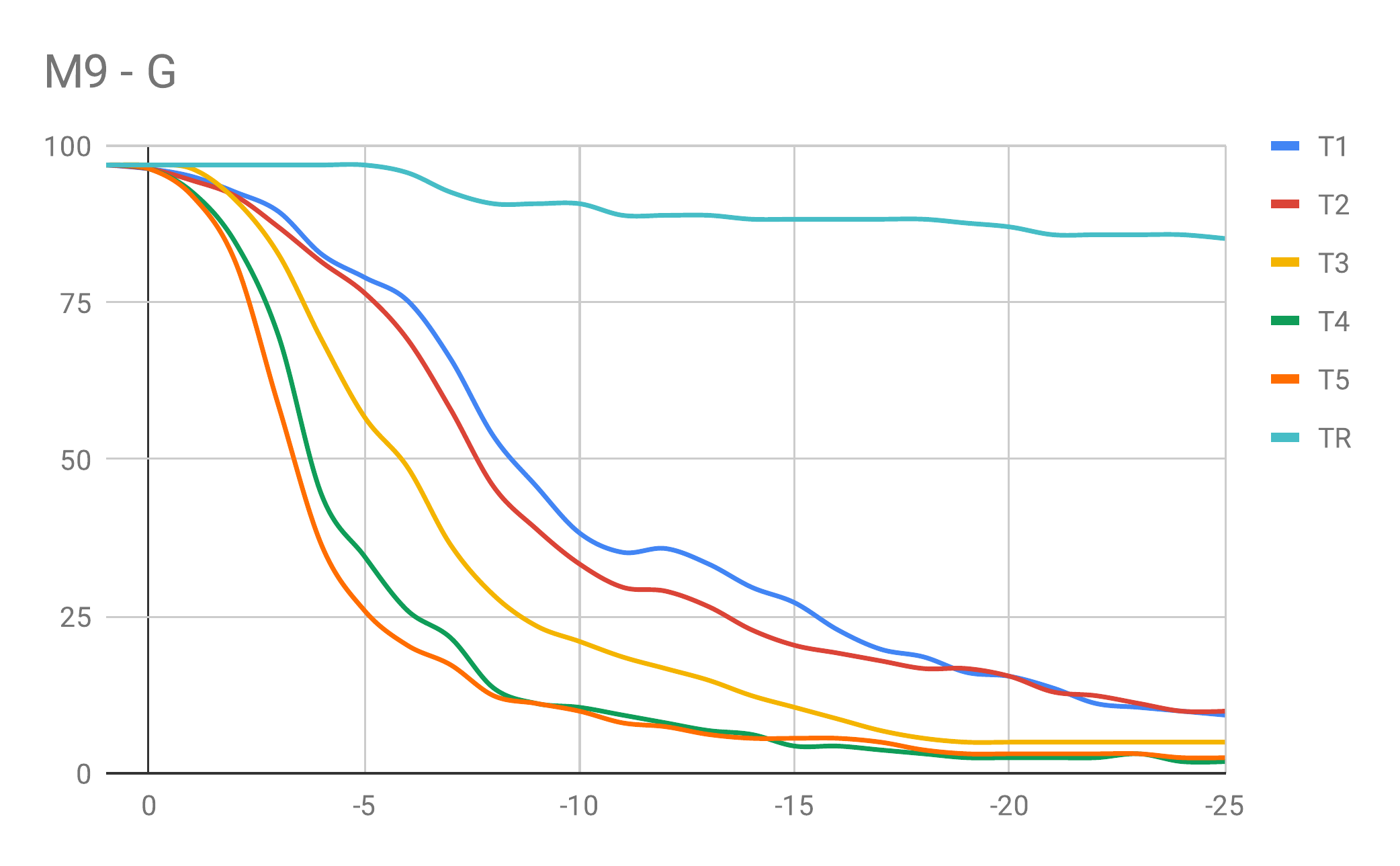}~~~\includegraphics[width=0.45\textwidth]{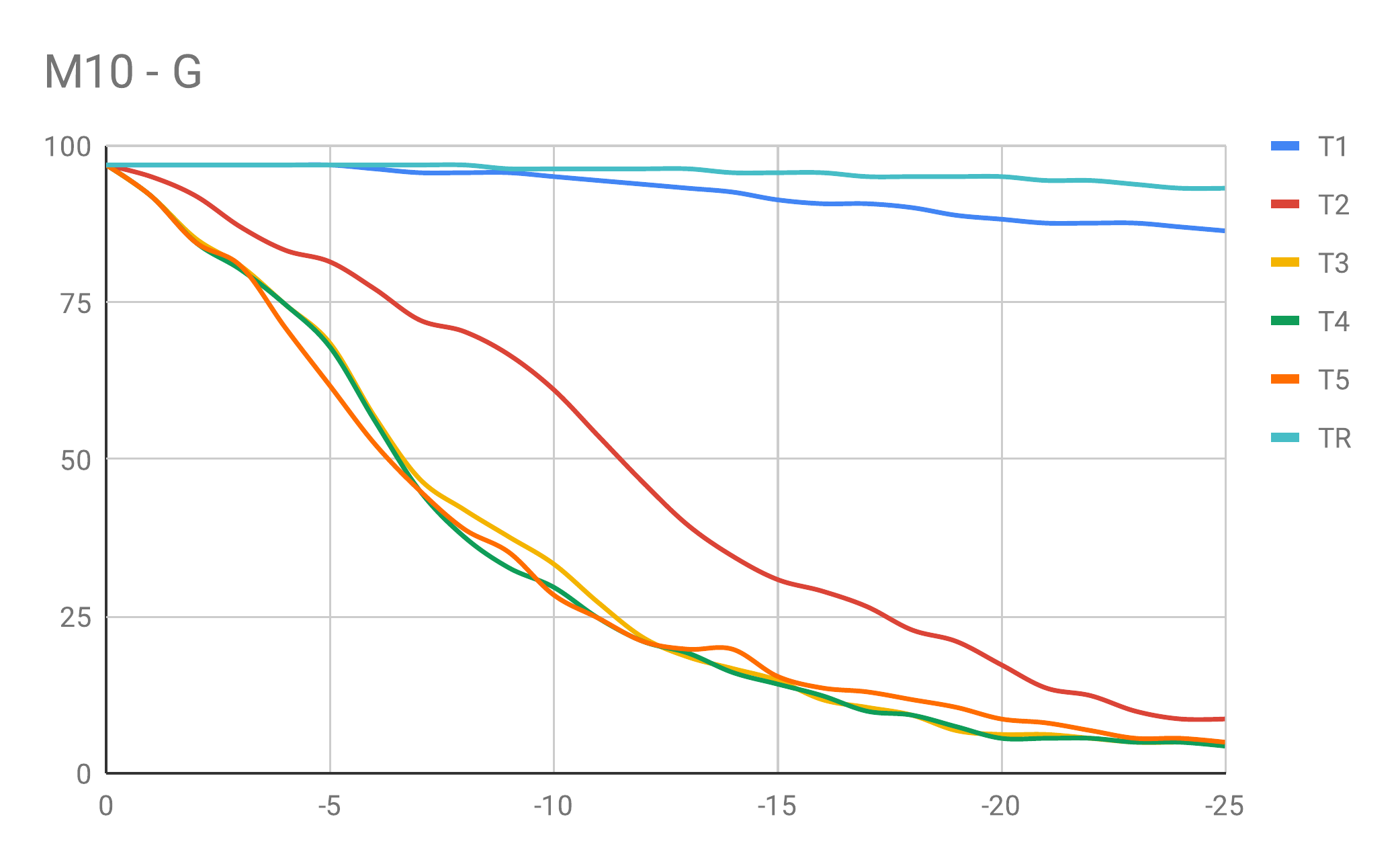}\\
\caption{Results when scaling the activations for the top 1-5 (T1 - T5) encoder dimensions associated with gradation for each of out ten models M1 - M10. These graphs show the amount of gold standard outputs undergoing gradation which are produced when encoder dimensions are scaled. As comparison, the green TR graph shows the effect of scaling 5 randomly selected encoder dimensions.}
\label{fig:all-g-scaling}
\end{figure*}

This appendix contains all results for  the scaling experiment presented in Section \ref{sec:experiments}. Figure \ref{fig:all-ng-scaling} presents the amount of alternate forms produced by each model when 1 - 5 top gradation encoding hidden state dimensions associated with gradation are scaled. Figure \ref{fig:all-g-scaling} presents results for the gold standard forms undergoing gradation. For each model, we also present results for scaling a set of five randomly selected encoder hidden state dimensions. 

As Figures \ref{fig:all-ng-scaling} and \ref{fig:all-g-scaling} show, the effect of scaling dimensions associated with gradation has a clear positive effect on the number of output forms which do not undergo gradation. In contrast, scaling randomly selected encode hidden state dimensions has small effect overall on the number of these output forms although it does tend to reduce the number of gold standard outputs undergoing gradation. This means that the number of nonce output forms still increases when the scaling factor approaches \(-25\) as might be expected because we are deviating from the learned models parameters.

\end{document}